\documentclass[11pt]{article}

\usepackage[final]{acl}

\usepackage{times}
\usepackage{latexsym}
\usepackage{threeparttable}

\usepackage[T1]{fontenc}
\usepackage[utf8]{inputenc}

\usepackage{microtype}

\usepackage{inconsolata}

\usepackage{graphicx}

\usepackage{booktabs}
\usepackage{makecell}
\usepackage{graphicx}
\usepackage{pifont}
\usepackage{booktabs}
\usepackage{tabularx}
\usepackage{makecell}

\newcommand{\cmark}{\ding{51}}
\newcommand{\xmark}{\ding{55}}
\usepackage{xcolor}
\usepackage{pifont}
\usepackage{tabularx}

\definecolor{softgreen}{HTML}{2E7D32}
\definecolor{softred}{HTML}{C62828}

\providecommand{\cmark}{}
\providecommand{\xmark}{}
\renewcommand{\cmark}{\textcolor{softgreen}{\ding{51}}}
\renewcommand{\xmark}{\textcolor{softred}{\ding{55}}}

\usepackage{amsmath}
\usepackage{amssymb}
\usepackage[table]{xcolor}
\usepackage{ifthen}
\usepackage{calc}
\usepackage{siunitx}
\makeatletter
\definecolor{softgreen}{RGB}{46,125,50}
\definecolor{softred}{RGB}{198,40,40}

\definecolor{colVGQ}{RGB}{90,156,219}     % calm sky-blue (trust/clarity)
\definecolor{colVEA}{RGB}{106,186,128}   % soft sage-green (harmony/accuracy)
\definecolor{colTGQ}{RGB}{238,201,120}    % muted warm amber (optimism/readability)
\definecolor{colTEA}{RGB}{233,134,126}   % soft coral (warmth/affect; not alarming)
\definecolor{colLC}{RGB}{173,148,214}     % gentle lavender (order/structure)
\definecolor{colHGQ}{RGB}{86,193,182}    % light teal (balance/refreshing)
\definecolor{colPPE}{RGB}{120,178,225}    % airy blue (confidence/consistency)
\definecolor{colPES}{RGB}{242,176,120}   % peach (approachability/positivity)
\definecolor{colTAS}{RGB}{164,176,189}     % cool light gray (neutral/technical)
\definecolor{colCFS}{RGB}{130,205,170}     % mint (ease/fluency)
\definecolor{colSSS}{RGB}{173,193,222}     % light periwinkle-gray (soothing, less heavy)
\definecolor{colRFS}{RGB}{210,120,120}% muted rose (summary emphasis, gentle)

\newcommand{\mspm}[2]{\ensuremath{#1_{\mbox{\tiny$\pm$#2}}}}
\newcommand{\model}[1]{\texttt{#1}}

\usepackage{xspace}
\makeatletter
\DeclareRobustCommand\onedot{\futurelet\@let@token\@onedot}
\def\@onedot{\ifx\@let@token.\else.\null\fi\xspace}

\makeatother

\usepackage[most]{tcolorbox}
\usepackage{xcolor}
\usepackage[most]{tcolorbox}
\usepackage{xcolor}
\usepackage{listings}

\newtcblisting{promptbox}[1][]{
  enhanced,
  breakable,
  listing only,
  listing engine=listings,
  colback=gray!3,
  colframe=black!50,
  colbacktitle=gray!60,
  coltitle=white,
  boxrule=0.4pt,
  arc=0pt,
  left=4pt,
  right=4pt,
  top=3pt,
  bottom=3pt,
  fonttitle=\bfseries\footnotesize,
  title={User Prompt},
  listing options={
    basicstyle=\scriptsize\rmfamily,
    breaklines=true,
    breakatwhitespace=false,
    breakindent=0pt,
    breakautoindent=false,
    postbreak={},
    columns=fullflexible,
    keepspaces=true,
    showstringspaces=false,
    tabsize=2
  },
  #1
}

\title{MER-Bench: Benchmarking Multimodal Emotional Reappraisal \\ for Image-Text Memes}

\author{
Yiqi Nie\textsuperscript{1,2}, Fei Wang\textsuperscript{2,}\thanks{Corresponding author.}, Junjie Chen\textsuperscript{3}, Kun Li\textsuperscript{4}, Yudi Cai\textsuperscript{2,5}, \\
\textbf{Dan Guo\textsuperscript{2,3}, Chenglong Li\textsuperscript{1}, Meng Wang\textsuperscript{2,3,$\ast$}}\\
\textsuperscript{1} Anhui University \\
\textsuperscript{2} Institute of Artificial Intelligence, Hefei Comprehensive National Science Center \\
\textsuperscript{3} Hefei University of Technology 
\textsuperscript{4} United Arab Emirates University \\
\textsuperscript{5} University of Science and Technology of China \\
{\small\texttt{nieyiqi5@gmail.com, wangfei@iai.ustc.edu.cn}} 
}

\begin{document}
\maketitle
\begin{abstract}
% Multimodal Emotional Reappraisal (MER) aims to shift affective interpretation by reinterpreting a situation without altering its underlying semantic context.
Multimodal Emotional Reappraisal (MER) aims to shift affective interpretation by rewriting and editing a situation without altering its underlying semantic context.
Despite advances in controllable editing and sentiment transfer, MER remains challenging under coupled multimodal constraints, especially in internet memes where images, embedded text, and layout jointly encode affective meaning.
To bridge this gap, we present \textbf{MER-Bench}, a systematic benchmark for MER on image-text memes, containing over 3,000 real-world memes with fine-grained annotations for emotion transformation and structure-preserving multimodal editing.
MER-Bench spans diverse visual, textual, and structural settings and is paired with a structured MLLM-as-a-Judge evaluation suite, comprising eight fine-grained metrics and three aggregated objectives for emotional alignment, content fidelity, and transformation strength.
Extensive evaluation of 14 advanced multimodal models reveals that:
\ding{182} current models struggle to jointly balance affect, semantics, structure, and text;
\ding{183} strong visual editing does not imply effective MER, with poor embedded text rendering and affect-aware rewriting remaining key bottlenecks.
Our results identify MER as a critical capability gap in current multimodal generation systems and highlight affective image-text reasoning as a key challenge for future work. Project Page: \href{https://github.com/one-seven17/MER-Bench}{https://github.com/one-seven17/MER-Bench}.

\end{abstract}

\addtocontents{toc}{\protect\setcounter{tocdepth}{-1}}

\section{Instruction}
\begin{figure}[t]
\centering
\includegraphics[width=1.0\columnwidth]{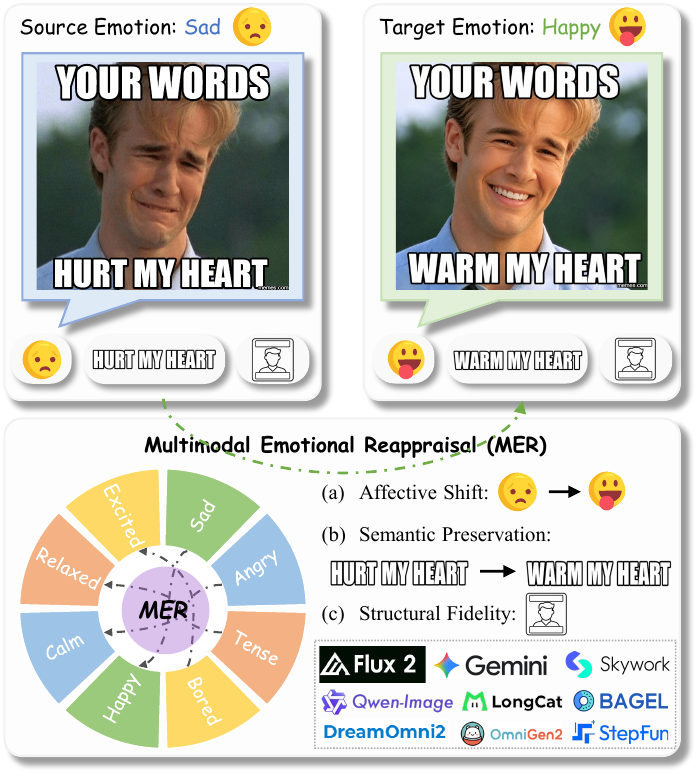}
\vspace{-1.5em}
\caption{
Illustration of MER, which requires affective shift, semantic preservation, and structural fidelity when transforming negative memes into positive ones.
}
\label{fig:dataset-showcase}
\vspace{-1.5em}
\end{figure}

Emotional Reappraisal is a fundamental cognitive strategy for shifting affective interpretation by reinterpreting a situation without altering its underlying semantic context~\cite{gross1998emerging,gross2014emotion,su2025context}.
As illustrated in Fig.~\ref{fig:dataset-showcase}, \textbf{M}ultimodal \textbf{E}motional \textbf{R}eappraisal (MER) extends this idea to image-text settings, requiring models to transform emotional framing while jointly preserving semantic intent and structural coherence.
Such reappraisal is particularly challenging in image-text memes, where affective meaning is jointly constructed through visual content, embedded text, layout structure, and culturally shared templates~\cite{mortensen2021playful,lin2026goat,denisova2019internet}.
As a prevalent form of online affective expression, memes frequently convey frustration, anxiety, sarcasm, ridicule, and polarized attitudes through highly compressed multimodal signals~\cite{brubaker2018one}, making them a challenging testbed for MER.

Recent studies~\cite{lin2026goat, sharma2024emotion, bai2025state, kiela2020hateful,yuan2025enhancing,ouyang2025sentiment,li2024mhrn} have substantially advanced meme understanding, harmful content detection, and template-aware generation~\cite{kiela2020hateful,lin2026goat,bai2025state,chen2024xmecap,cai2025itermeme,wang2024frequency,wang2024eulermormer}.
Meanwhile, multimodal generation systems have demonstrated strong capabilities in controllable editing and affect-aware generation~\cite{yang2025emoctrl,yang2024emogen,yang2025emoedit,wang2026xinsight}.
However, existing approaches primarily focus on recognition, generation, or coarse sentiment control, leaving emotional reappraisal under coupled multimodal constraints largely unexplored.
Compared with conventional sentiment editing, MER introduces several unique challenges:
(1) emotional meaning in memes is often implicit, culturally grounded, and structurally entangled across modalities, requiring multimodal reasoning over both explicit and implicit cues~\cite{wiggins2019discursive};
(2) emotional transformation requires controllable modeling of directional affect shifts while preserving multimodal semantic consistency;
and (3) evaluation is inherently multi-objective, since valid outputs must jointly satisfy emotional transformation, structural preservation, and content fidelity.

Towards this end, we present \textbf{\textit{MER-Bench}}, a benchmark for MER on image-text memes.
Grounded in cognitive reappraisal theory and Russell's circumplex model of affect, MER-Bench operationalizes emotional reappraisal through four emotion-shift pairs, structured prompt-based rewriting protocols, and structure-preserving multimodal editing objectives.
The benchmark contains over 3,000 real-world memes spanning diverse visual modalities, structural layouts, and affective conditions, with fine-grained annotations covering emotion transition, meme text rewriting, visual editing specifications, and structural taxonomy labels.
These annotations enable systematic evaluation of whether models can achieve controllable affective shifts while preserving the semantic and structural identity of the original meme.

To systematically evaluate model capabilities under coupled multimodal constraints, we further introduce a structured \textbf{MLLM-as-a-Judge} evaluation suite.
The framework includes eight fine-grained metrics:
Visual Generation Quality (VGQ),
Visual Emotion Alignment (VEA),
Text Generation Quality (TGQ),
Text Emotion Alignment (TEA),
Layout Consistency (LC),
Holistic Generation Quality (HGQ),
Perceived Primary Emotion (PPE) and Perceived Emotion Shift (PES).
To further capture the conjunctive nature of emotional reappraisal, we additionally introduce three aggregated objectives:
Target Alignment Score (TAS),
Content Fidelity Score (CFS), and
Shift Strength Score (SSS),
which are unified into the final Reappraisal Fidelity Score (RFS).
To validate the reliability of the proposed evaluation protocol, we conduct human evaluation and cross-configuration analysis under different prompting strategies.
Four human experts score sampled outputs across multiple prompting configurations, providing reference annotations for cross-validation with MLLM-based judgments.
The strong agreement across multiple statistical measures confirms the effectiveness, stability, and human alignment of our automatic evaluation framework.
Applying this evaluation suite to 14 advanced multimodal generation models, we uncover two key findings:
\ding{182} Current models often satisfy only part of MER's coupled requirements, struggling to jointly preserve structure, maintain semantic identity, and perform target affective reinterpretation;
\ding{183} MER is a multimodal reasoning problem rather than a visual editing or positive-generation task, as models must preserve the structural and pragmatic conditions of meme meaning while modifying affective framing.

% \ding{182} Emotional reappraisal requires jointly preserving meme structure, semantic identity, and affective reinterpretation, whereas current models typically satisfy only part of these coupled constraints.
% \ding{183} MER is a multimodal reasoning problem rather than a visual editing task. Category-level analysis shows that reappraisal difficulty increases with structural rigidity and narrative complexity: template-based and multi-panel memes are harder to reinterpret, while positive-polarity memes are easier for current models to transform.
Thus, our contributions are as follows:
\begin{itemize}
% \item \textbf{\textit{Systematic Benchmark}}: We introduce MER-Bench, a novel benchmark for MER in image-text memes, organized around three taxonomic axes and nine subcategories to evaluate controllable emotional shifts and structure-preserving multimodal editing.
\item \textbf{\textit{Systematic Benchmark}}: We introduce MER-Bench, a novel benchmark for MER in image-text memes, organized around three taxonomic axes and nine subcategories to evaluate controllable affective shifts under structure-preserving multimodal constraints.
% \item \textbf{\textit{Conjunctive Evaluation Protocol}}: We introduce a structured MLLM-as-a-Judge evaluation suite to assess MER under coupled constraints. In addition, the effectiveness of our evaluation strategy has been thoroughly validated by four human experts.
% \item \textbf{\textit{Evaluation Suite}}: We propose a structured MLLM-as-a-Judge evaluation suite for assessing MER under coupled multimodal constraints, and validate its reliability through human expert study and prompt ablation.
\item \textbf{\textit{Evaluation Suite}}: We propose a structured MLLM-as-a-Judge evaluation suite for assessing MER under coupled multimodal constraints, and validate its reliability through human expert study and prompt ablation.
% \item \textbf{\textit{Empirical Analysis}}: We evaluate 14 advanced multimodal models on MER-Bench, revealing their strengths and limitations and establishing a foundation for future research on MER.
\item \textbf{\textit{Empirical Analysis}}: We evaluate 14 advanced multimodal models on MER-Bench, showing that effective MER requires coordinated affective transformation, semantic preservation, structural consistency, and embedded text rewriting, especially when meme meaning depends on rigid templates, sequential layouts, or ambiguous affective cues.
\end{itemize}

\section{Related Works}
% \begin{table}
% \centering
% \setlength{\tabcolsep}{2pt}
% \resizebox{\linewidth}{!}{
% \begin{threeparttable}
% \caption{
% Benchmark comparison across emotion-related multimodal tasks
% }
% \label{tab:emotion_benchmarks}
% \begin{tabular}{
% l|ccccc
% }
% \hline
% \textbf{Dataset} &
% \textbf{Modality} &
% \makecell{\textbf{Emotion}\\\textbf{Recognition}} &
% \makecell{\textbf{Emotion Reasoning}\\ \textbf{/ Regulation}} &
% \makecell{\textbf{Emotion}\\\textbf{Generation}} &
% \makecell{\textbf{Social}\\\textbf{Media}} \\
% \hline
% Memotion~3~\cite{mishra2023memotion}       & Image-Text  & \cmark & \xmark & \xmark & \cmark \\
% MOOD~\cite{sharma2024emotion}              & Image-Text  & \cmark & \xmark & \xmark & \cmark \\
% GOAT-Bench~\cite{lin2024goat}              & Image-Text  & \cmark & \xmark & \xmark & \cmark \\
% EmotionBench~\cite{huang2023emotionally}   & Text         & \cmark & \cmark & \cmark & \xmark \\
% EmoBench~\cite{sabour2024emobench}         & Text         & \cmark & \cmark & \cmark & \xmark \\
% MOSABench~\cite{song2024mosabench}         & Image        & \cmark & \xmark & \xmark & \xmark \\
% MM-InstructEval~\cite{yang2025mm}          & Image        & \cmark & \xmark & \cmark & \xmark \\ \hline
% MER-Bench (Ours)                             & Image-Text  & \cmark & \cmark & \cmark & \cmark \\
% \hline
% \end{tabular}
% \end{threeparttable}
% }
% \vspace{-1.5em}
% \end{table}

\begin{table*}[t]
\centering
\small
\setlength{\tabcolsep}{10pt}
\caption{Benchmark comparison across emotion-related multimodal tasks.}
\label{tab:emotion_benchmarks}
\resizebox{1.0\textwidth}{!}{%
\begin{tabular}{|l|l|cccc|}
\hline
{Dataset} &
{Modality} &
\makecell{{Emotion}\\{Recognition}} &
\makecell{{Emotion Reasoning}\\{/ Regulation}} &
\makecell{{Emotion}\\{Generation}} &
\makecell{{Social}\\{Media}} \\
\hline
Memotion~3~\cite{mishra2023memotion}     & Image-Text & \cmark & \xmark & \xmark & \cmark \\
MOOD~\cite{sharma2024emotion}            & Image-Text & \cmark & \xmark & \xmark & \cmark \\
GOAT-Bench~\cite{lin2026goat}            & Image-Text & \cmark & \xmark & \xmark & \cmark \\
EmotionBench~\cite{huang2023emotionally} & Text        & \cmark & \cmark & \cmark & \xmark \\
EmoBench~\cite{sabour2024emobench}       & Text        & \cmark & \cmark & \cmark & \xmark \\
MOSABench~\cite{song2024mosabench}       & Image       & \cmark & \xmark & \xmark & \xmark \\
MM-InstructEval~\cite{yang2025mm}        & Image       & \cmark & \xmark & \cmark & \xmark \\
\hline
\textbf{MER-Bench (Ours)}                & \textbf{Image-Text} & \cmark & \cmark & \cmark & \cmark \\
\hline
\end{tabular}%
}
\end{table*}

\subsection{Meme Emotion Understanding}
Early benchmarks, such as the Facebook Hateful Memes challenge, formulate meme analysis as multimodal hate speech detection, advancing image-text fusion and robustness to spurious correlations~\cite{kiela2020hateful}. 
Subsequent datasets, including MemeInterpret~\cite{park2025memeinterpret} and GOAT-Bench~\cite{lin2026goat}, introduce culturally grounded categories and fine-grained labels for real-world distribution shifts. 
Beyond classification, MemeReaCon~\cite{zhao2025memereacon} and MET-Meme~\cite{xu2022met} require reasoning over social context, user comments, communicative intent, and multimodal metaphors. 
Sharma introduces MOOD, a six-emotion meme dataset, and ALFRED, an emotion-aware multimodal fusion model using visual cues and gated cross-modal fusion~\cite{sharma2024emotion}.
Memes have also been studied as socio-psychological signals of collective attitudes and public discourse~\cite{brubaker2018one,wang2026multimodal,wang2026xinsight}.

\subsection{Meme Generation}
Meme generation spans template-aware captioning, text-to-image generation, and punchline generation.
Layout-aware systems XMeCap~\cite{chen2024xmecap} and IterMeme~\cite{cai2025itermeme} jointly model template structures and humorous text via template-conditioned or cloze-style generation.
MemeBot~\cite{sadasivam2020memebot} frames meme creation as a translation-like process combining template images with captions.
Large-scale resources such as Memeify~\cite{vyalla2020memeify} support caption generation, while MemeCraft~\cite{wang2024memecraft} leverages LLMs/VLMs for socially aligned meme synthesis.
HUMOR adopts multi-path chain-of-thought reasoning with preference-based reinforcement learning~\cite{li2025perception}, while theory-guided frameworks~\cite{zhang2026humorchain,li2025mmad} integrate psychological humor theories.

% However, existing systems focus on generating new memes or captions, without explicitly modeling affect transformation under structural-preservation constraints.

\subsection{Emotion-Controllable Generation}
Emotion-controllable generation has been explored in both textual and visual domains. Text-based methods perform attribute-conditioned style transfer while approximately preserving semantic content~\cite{xu2025prompt,li2026ma}, with extensions addressing finer cues such as sarcasm. 
% Visual approaches manipulate affect through expression editing or prompt-guided conditioning, as in EmoGen~\cite{yang2024emogen} and EmoCtrl~\cite{yang2025emoctrl}.
Visual approaches control affect through emotion-conditioned generation or image manipulation, as in EmoGen~\cite{yang2024emogen}, EmoCtrl~\cite{yang2025emoctrl}, and EmoEdit~\cite{yang2025emoedit}. 
Diffusion-based models such as T2i-Adapter~\cite{mou2024t2i}, IP-Adapter~\cite{ye2023ip}, and Instruct-Pix2Pix~\cite{brooks2023instructpix2pix} enable structured conditioning for layout and style control. Diffusion-DPO~\cite{wallace2024diffusion} instead aligns diffusion models with human preferences through direct preference optimization.
% As summarized in Table~\ref{tab:emotion_benchmarks}, existing work typically targets generic sentences or images and models coarse emotional dimensions. 
% None explicitly evaluates structure-preserving affect transformation in meme settings, where models must jointly maintain visual-verbal structure, scenario integrity, and psychologically grounded reappraisal patterns.
As summarized in Tab.~\ref{tab:emotion_benchmarks}, existing methods focus on coarse affect control and cannot jointly satisfy the semantic, structural, and reappraisal constraints required for meme transformation.

\begin{figure*}[!t]
\centering
\includegraphics[width=\textwidth]{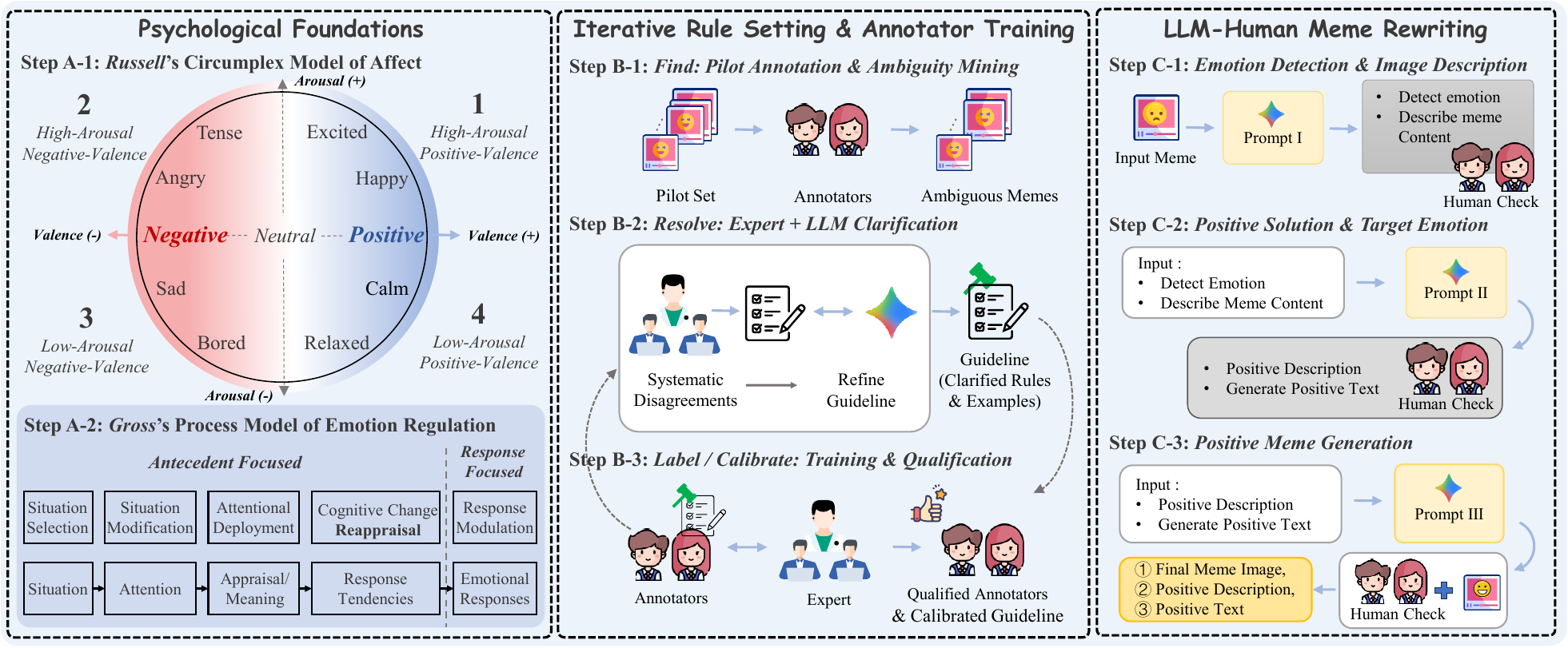}
\vspace{-1.5em}
% \caption{\textbf{Construction pipeline of MER-Bench.}
% (\textbf{A}) \textit{Psychological foundations} define the affective space using Russell's circumplex model and ground the task in cognitive reappraisal theory.
% (\textbf{B}) \textit{Iterative rule setting and annotator training} follow a Find--Resolve--Label paradigm, where pilot annotation, expert clarification, and calibration are used to refine guidelines.
% (\textbf{C}) \textit{LLM-assisted human rewriting} uses a three-stage workflow for emotion detection, target reappraisal specification, and final meme generation, followed by human verification.
\caption{\textbf{Construction pipeline of MER-Bench.}
(\textbf{A}) Psychological theories define the affective space and reappraisal objective for emotion transformation.
(\textbf{B}) Iterative annotation, expert clarification, and annotator calibration refine labeling guidelines and resolve ambiguous cases.
(\textbf{C}) LLM-assisted human rewriting performs emotion detection, reappraisal planning, and structure-preserving image-text meme generation with human verification.
}
\label{fig:data-construction}
\vspace{-1em}
\end{figure*}

\section{Task Definition}
\label{sec:task}
Let $\mathcal{M}=(I,T,\ell)$ denote an image-text meme, where $I$, $T$, and $\ell$ represent the visual content, overlaid text, and layout structure, respectively.
We formulate MER as a constrained generation task grounded in emotion regulation theory~\cite{gross1998emerging,gross2014emotion}.
Given a source meme $\mathcal{M}^{-}=(I^{-},T^{-},\ell^{-})$ with source affect $a^{-}$ and a target affect $a^{+}$, the goal is to generate a reappraised meme $\mathcal{M}^{+}=(I^{+},T^{+},\ell^{+})$ that shifts the perceived affect from $a^{-}$ toward $a^{+}$ while preserving the original semantic situation and structure:
\begin{equation}
\mathcal{M}^{+} = f(\mathcal{M}^{-}, a^{+}).
\end{equation}

The generated meme should satisfy three coupled requirements:
\begin{equation}
a(\mathcal{M}^{+}) \approx a^{+}, \quad
S(\mathcal{M}^{+}) \approx S(\mathcal{M}^{-}), \quad
\ell^{+} \approx \ell^{-},
\end{equation}
where $a(\cdot)$ denotes perceived affective framing, and $S(\cdot)$ denotes the underlying semantic situation, including core entities, events, and relations.
Unlike conventional sentiment editing or open-ended meme generation, MER requires changing how a meme is emotionally interpreted without changing what it is about or how its multimodal content is structurally organized.
Thus, successful MER requires jointly satisfying affective shift, semantic preservation, and structural fidelity.

\section{Dataset Construction}
\label{sec:dataset}

To benchmark MER in a controlled and comparable setting, we construct \textbf{\textit{MER-Bench}} from the \textit{MOOD} meme corpus~\cite{sharma2024emotion}.
Starting with 10,005 memes, we retain 6,695 samples with clearly identifiable negative affective framing and then apply quality, duplication, safety, and reappraisal-feasibility filters, resulting in 3,117 high-quality image-text meme instances.
Detailed filtering statistics are provided in Appendix~\ref{app:dataset-details}.
Fig.~\ref{fig:data-construction} illustrates the overall construction pipeline.

\subsection{Psychological Foundations}
We define the reappraisal target space using Russell's circumplex model of affect~\cite{russell1980circumplex}.
As shown in Fig.~\ref{fig:data-construction} (A-1), MER-Bench covers four negative-to-constructive affective transitions:
\textit{Sad} $\rightarrow$ \textit{Happy},
\textit{Angry} $\rightarrow$ \textit{Calm},
\textit{Tense} $\rightarrow$ \textit{Relaxed}, and
\textit{Bored} $\rightarrow$ \textit{Excited},
with 1,026, 937, 1,020, and 134 samples (Fig.~\ref{fig:data_appendix}), respectively.
Following Gross's process model of emotion regulation~\cite{gross1998emerging,gross2014emotion}, we design a reappraisal-oriented rewriting protocol that changes affective interpretation while preserving the underlying semantic situation and structure.

\subsection{Construction Pipeline and Control}
MER-Bench is constructed through an LLM-assisted human annotation workflow.
Five trained annotators with graduate-level backgrounds participate in annotation, while four experts, including three computer science experts and one psychology expert, support rule refinement, ambiguity resolution, and quality control.
Each candidate sample is checked by four annotators, with unresolved disagreements escalated to expert review.

After annotator calibration, we use a three-stage prompting workflow:
(i) emotion detection and image description,
(ii) reappraisal planning with target emotion and editing intent, and
(iii) reappraised meme generation.
The first two stages use \texttt{Gemini-2.5-Flash}\footnote{\url{https://deepmind.google/technologies/gemini/flash/}}, while the final multimodal generation stage uses \texttt{Gemini-2.5-Flash-Image}\footnote{\url{https://ai.google.dev/gemini-api/docs/models/gemini-2.5-flash-image}}.
LLMs are used only as annotation assistants; all retained samples are manually verified before inclusion.
Detailed prompt templates are provided in Appendix~\ref{app:prompt-templates}.
Each sample is checked for source-emotion validity, target-emotion consistency, semantic preservation, and structural editability.
Samples are revised or removed if the source affect is ambiguous, the target reappraisal is infeasible, the reappraised caption changes the core situation, or the generated meme fails to preserve essential layout and entity information.
The complete annotation schema, human-control statistics, and inter-annotator agreement are provided in Appendix~\ref{app:dataset-details}.

\subsection{Dataset Statistics}
\label{sec:dataset-statistics}
\begin{figure}[!t]
\centering
\includegraphics[width=\columnwidth]{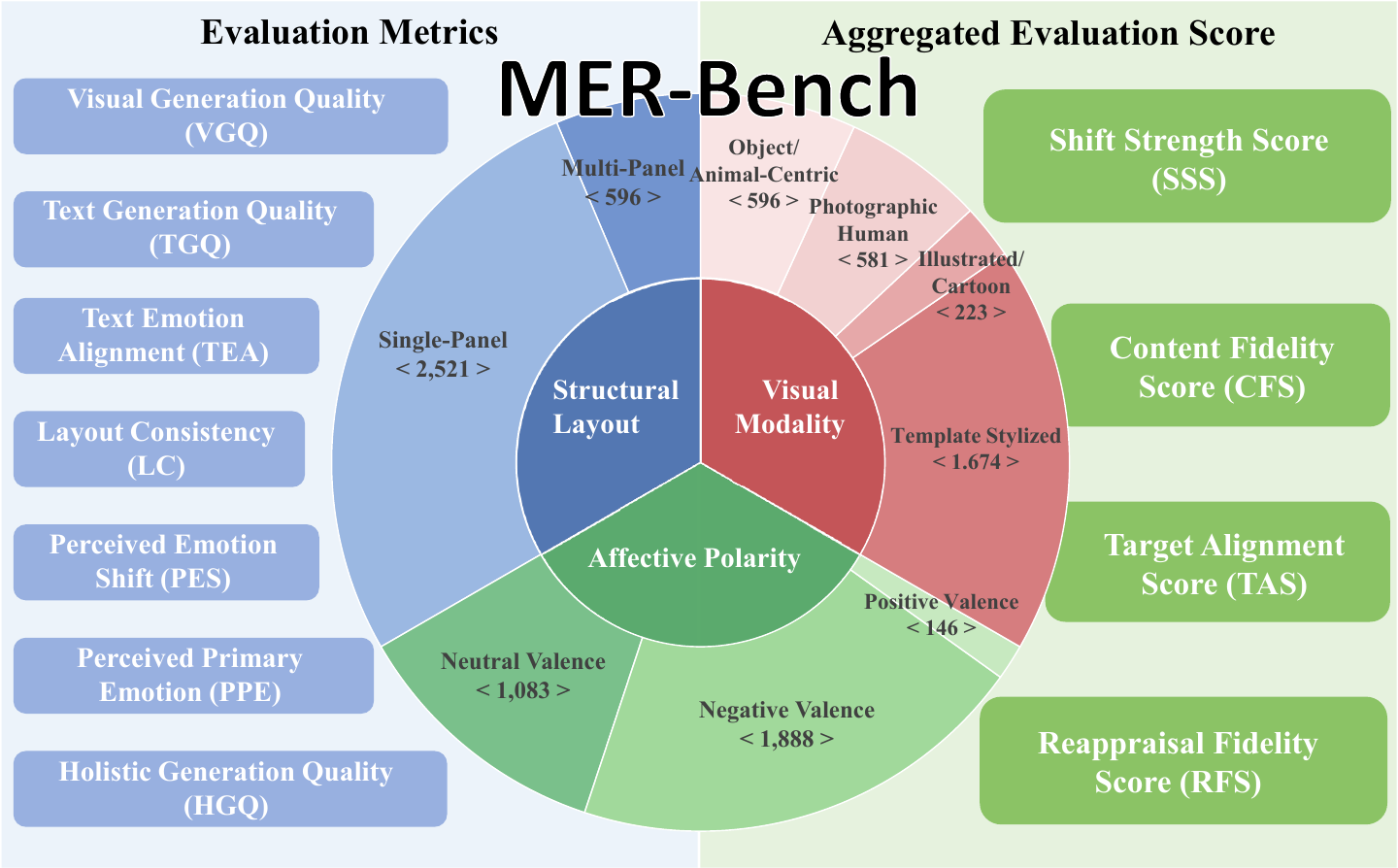}
\caption{
MER-Bench provides a unified taxonomy and evaluation metrics for MER.
}
\vspace{-1.5em}
\label{fig:dataset-stat}
\end{figure}

Fig.~\ref{fig:dataset-stat} summarizes the distribution of MER-Bench across three taxonomy axes and nine subcategories.
The visual modality axis includes four categories: object/animal-centric, photographic human, illustrated/cartoon, and template-stylized memes.
The structural layout axis distinguishes single-panel and multi-panel memes.
The affective polarity axis describes the caption-level sentiment polarity of the original meme text, including negative, neutral, and positive captions.
This taxonomy characterizes MER difficulty from complementary perspectives: visual grounding, layout complexity, and textual affective signal.
Template-stylized memes form the largest visual category, reflecting the prevalence of template-based meme creation, while single-panel memes dominate the structural distribution, matching common meme formats on social media.
Although MER-Bench is constructed with a negative holistic affective framing, some source captions are neutral or positive on the surface because negative affect can also be conveyed through visual context, layout, irony, or image-text incongruity.

\section{Evaluation Protocol}
\label{sec:evaluation-protocol}
\subsection{MLLM-as-a-Judge}
\label{subsec:mllm-as-a-judge}
Evaluating MER is inherently multi-objective: a valid output must express the target emotion while preserving semantic content, meme structure, and textual readability.
Single-axis metrics such as sentiment accuracy or semantic similarity cannot capture these coupled requirements~\cite{sharma2024emotion,sharma2026memetag,lin2026goat}.
We adopt a structured \emph{MLLM-as-a-Judge} paradigm~\cite{pmlr-v235-chen24h}, using \texttt{Gemini-3-Pro}\footnote{\url{https://deepmind.google/models/gemini/pro/}} as the judge.
Given the source meme, edited meme, and target emotion, the judge returns a schema-constrained JSON output.
To reduce free-form judgment bias, we impose three constraints:
(i) \emph{target-gated emotion scoring}, which limits alignment scores when the perceived emotion mismatches the target;
(ii) \emph{localized rationales}, which justify each score using criterion-specific evidence;
and (iii) \emph{explicit layout verification}, which checks panel structure, caption placement, and spatial consistency.
The complete judge prompt and JSON schema are provided in Appendix~\ref{app:judge-prompt}.

We define eight fine-grained metrics covering visual quality, text quality, emotional alignment, structural fidelity, and holistic reappraisal success.
Except for \textbf{\emph{Perceived Primary Emotion (PPE)}}, which records the dominant emotion from an eight-emotion label space plus \texttt{other}, all scalar metrics use a 1-5 Likert scale~\cite{mellor2014use}:
\textbf{\emph{Visual Generation Quality (VGQ)}} assesses image quality and artifacts;
\textbf{\emph{Visual Emotion Alignment (VEA)}} measures visual match to the target emotion;
\textbf{\emph{Text Generation Quality (TGQ)}} evaluates caption legibility, formatting, fluency, and meme-style naturalness;
\textbf{\emph{Text Emotion Alignment (TEA)}} measures whether the caption conveys the target emotion while preserving the scenario;
\textbf{\emph{Layout Consistency (LC)}} checks panel structure, caption placement, and spatial organization;
\textbf{\emph{Holistic Generation Quality (HGQ)}} assesses overall visual-text coherence, readability, and content preservation;
and \textbf{\emph{Perceived Emotion Shift (PES)}} quantifies the strength of the target-consistent emotional shift.
Together, these metrics disentangle quality, controllability, structure, and overall fidelity of reappraisal.
\subsection{Prompt Sensitivity and Human Validation}
\label{subsec:prompt-validation}

Prompt design can affect the stability of MLLM-based evaluation.
We study two orthogonal prompt factors: evidence order and reasoning order.
For evidence order, we compare \textbf{\emph{Visual-First (VF)}} and \textbf{\emph{Visual-Last (VL)}} settings, which place visual evidence before or after textual task information.
For reasoning order, we compare \textbf{\emph{Rationale-First (RF)}} and \textbf{\emph{Rationale-Last (RL)}} settings, which require localized rationales before or after scalar scores.
Their combination yields four judge configurations: RF-VF, RF-VL, RL-VF, and RL-VL.
Following expert-based affective evaluation practice in MME-Emotion~\cite{zhang2026mme}, we conduct a human validation study to assess the reliability of our MLLM-as-a-Judge protocol.
Four experts manually score 100 randomly sampled MER outputs using the same evaluation rubric as the automatic judge.
These expert annotations serve as human reference scores for evaluating judge alignment across different prompt configurations.

\begin{figure}
\centering
\includegraphics[width=\linewidth]{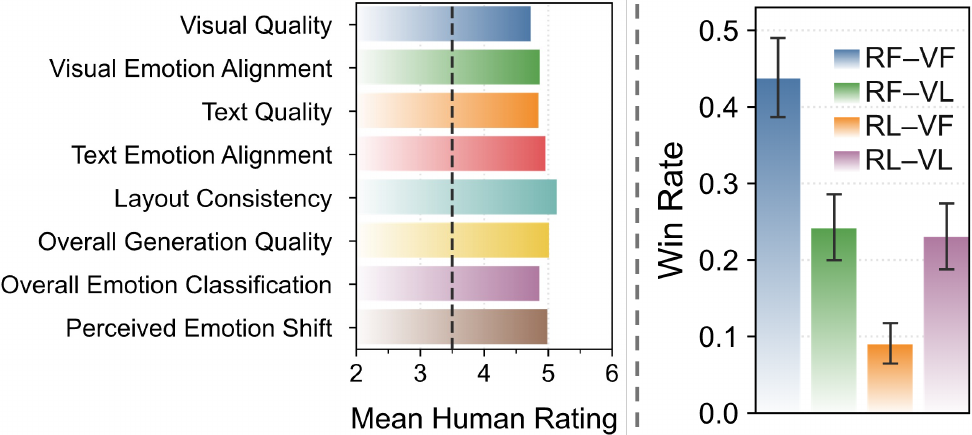}
% \vspace{-2.0em}
\caption{
Human validation and prompt ablation results for the MLLM-as-a-Judge evaluation protocol on 100 randomly sampled MER outputs. 
% Left: Mean human ratings (1--6 scale) across evaluation dimensions. The vertical dashed line indicates the neutral midpoint (score = 3.5), showing that all dimensions are rated positively above neutrality. 
% Right: Win rate comparison of four prompt configurations combining rationale order (RF \vs \ RL) and visual order (VF \vs VL). 
% The Rationale-First, Visual-First (RF-VF) setting achieves the highest win rate, indicating the strongest alignment with human preference.
}
\label{fig:eval-setting}
\vspace{-1.5em}
\end{figure}

We compare MLLM-based judgments with expert scores to examine whether the automatic evaluation captures human perception of multimodal quality, emotional alignment, and reappraisal success.
In Fig.~\ref{fig:eval-setting}, RF-VF achieves the highest agreement with human judgments among the four configurations.
This suggests that presenting visual evidence first and eliciting localized rationales before scalar scores better aligns MLLM-based evaluation with expert assessment.
Overall, the results support for the reliability and human alignment of our schema-constrained MLLM-as-a-Judge protocol.
% \subsection{Prompt Design and Human Validation}
% Prompt design critically affects the stability of MLLM-based evaluation.
% We investigate two orthogonal prompt factors: evidence order, including \textbf{\emph{Visual-First (VF)}} and \textbf{\emph{Visual-Last (VL)}}, and reasoning order, including \textbf{\emph{Rationale-First (RF)}} and \textbf{\emph{Rationale-Last (RL)}}.
% Their combination yields four configurations: RF-VF, RF-VL, RL-VF, and RL-VL.
% Following the expert-based affective evaluation practice adopted in MME-Emotion~\cite{zhang2025mme}, we further conduct a human validation study to assess the reliability of our MLLM-as-a-Judge protocol.
% Specifically, four experts manually score 100 randomly sampled Emotional Reappraisal outputs using the same evaluation dimensions as the automatic judge, providing human reference annotations for cross-validation.
% We then compare expert annotations with MLLM-based judgments across different prompting configurations to examine whether the automatic evaluation aligns with human perception of multimodal quality and affective transformation.Fig.~\ref{fig:eval-setting} shows that RF-VF achieves the highest agreement with human judgments, suggesting that visual-first evidence presentation and rationale-first scoring better align automated evaluation with human perception.
% These results confirm that our schema-constrained MLLM-as-a-Judge protocol is stable, interpretable, and human-aligned.

\begin{table*}[t]
\setlength{\tabcolsep}{3.2pt}
\renewcommand{\arraystretch}{1.18}
\centering
\caption{Model performance across evaluation metrics}
\label{tab:main_results_sorted_heat}
\vspace{-0.6em}
\resizebox{\textwidth}{!}{%
\begin{threeparttable}
\begin{tabular}{|l|cccccccc|ccc||c|}
\hline
% \textbf{Model} & \multicolumn{2}{c|}{Visual} & \multicolumn{2}{c|}{Textual}  & Layout & Holistic & \multicolumn{2}{c|}{Emotion} & & & &  \\ 
\textbf{Model} & 
\textbf{VGQ} $\uparrow$& 
\textbf{VEA} $\uparrow$& 
\textbf{TGQ} $\uparrow$& 
\textbf{TEA} $\uparrow$& 
\textbf{LC} $\uparrow$& 
\textbf{HGQ} $\uparrow$& 
\textbf{PPE} $\uparrow$& 
\textbf{PES} $\uparrow$& 
\textbf{TAS} $\uparrow$& 
\textbf{CFS} $\uparrow$& 
\textbf{SSS} $\uparrow$& 
\textbf{RFS} $\uparrow$\\
\hline

Flux9B~\cite{flux-2-2025} &
\cellcolor{colVGQ!82}\mspm{91.10}{1.16} &
\cellcolor{colVEA!100}\mspm{98.12}{0.98} &
\cellcolor{colTGQ!73}\mspm{58.27}{2.00} &
\cellcolor{colTEA!100}\mspm{92.47}{1.57} &
\cellcolor{colLC!78}\mspm{87.90}{2.33} &
\cellcolor{colHGQ!85}\mspm{68.77}{1.61} &
\cellcolor{colPPE!88}\mspm{79.50}{2.99} &
\cellcolor{colPES!100}\mspm{99.70}{0.28} &
\cellcolor{colTAS!85}\mspm{75.70}{3.07} &
\cellcolor{colCFS!85}\mspm{62.81}{1.61} &
\cellcolor{colSSS!100}\mspm{94.98}{0.04} &
\cellcolor{colRFS!100}\mspm{76.78}{1.16} \\

QwenEdit~\cite{wu2025qwenimagetechnicalreport} &
\cellcolor{colVGQ!73}\mspm{84.08}{2.20} &
\cellcolor{colVEA!93}\mspm{91.70}{2.33} &
\cellcolor{colTGQ!65}\mspm{50.68}{4.01} &
\cellcolor{colTEA!77}\mspm{80.92}{2.57} &
\cellcolor{colLC!76}\mspm{85.80}{3.33} &
\cellcolor{colHGQ!78}\mspm{61.25}{3.21} &
\cellcolor{colPPE!84}\mspm{75.10}{3.03} &
\cellcolor{colPES!96}\mspm{94.75}{2.06} &
\cellcolor{colTAS!74}\mspm{64.57}{3.30} &
\cellcolor{colCFS!75}\mspm{54.84}{3.99} &
\cellcolor{colSSS!99}\mspm{94.16}{0.35} &
\cellcolor{colRFS!90}\mspm{69.38}{2.77} \\

Flux4B~\cite{flux-2-2025} &
\cellcolor{colVGQ!70}\mspm{78.50}{1.53} &
\cellcolor{colVEA!96}\mspm{95.98}{1.82} &
\cellcolor{colTGQ!39}\mspm{25.37}{1.49} &
\cellcolor{colTEA!71}\mspm{74.42}{2.22} &
\cellcolor{colLC!75}\mspm{84.60}{4.03} &
\cellcolor{colHGQ!64}\mspm{47.35}{1.47} &
\cellcolor{colPPE!83}\mspm{74.30}{4.32} &
\cellcolor{colPES!99}\mspm{98.85}{0.76} &
\cellcolor{colTAS!72}\mspm{62.29}{4.05} &
\cellcolor{colCFS!58}\mspm{38.49}{2.05} &
\cellcolor{colSSS!99}\mspm{94.85}{0.12} &
\cellcolor{colRFS!82}\mspm{61.06}{1.75} \\

UniPic~\cite{wei2026skywork} &
\cellcolor{colVGQ!67}\mspm{74.78}{1.24} &
\cellcolor{colVEA!57}\mspm{61.72}{2.92} &
\cellcolor{colTGQ!58}\mspm{42.88}{3.35} &
\cellcolor{colTEA!62}\mspm{62.85}{4.39} &
\cellcolor{colLC!72}\mspm{81.50}{1.72} &
\cellcolor{colHGQ!55}\mspm{40.27}{2.58} &
\cellcolor{colPPE!59}\mspm{50.50}{4.74} &
\cellcolor{colPES!63}\mspm{70.13}{2.92} &
\cellcolor{colTAS!44}\mspm{31.45}{3.76} &
\cellcolor{colCFS!61}\mspm{41.17}{2.09} &
\cellcolor{colSSS!88}\mspm{87.76}{1.10} &
\cellcolor{colRFS!63}\mspm{48.43}{2.48} \\

DreamO2~\cite{xia2025dreamomni2} &
\cellcolor{colVGQ!60}\mspm{68.47}{2.54} &
\cellcolor{colVEA!46}\mspm{46.25}{4.77} &
\cellcolor{colTGQ!42}\mspm{27.80}{2.70} &
\cellcolor{colTEA!51}\mspm{48.30}{3.50} &
\cellcolor{colLC!79}\mspm{89.00}{2.94} &
\cellcolor{colHGQ!41}\mspm{25.45}{2.39} &
\cellcolor{colPPE!51}\mspm{43.60}{4.84} &
\cellcolor{colPES!52}\mspm{55.45}{4.99} &
\cellcolor{colTAS!34}\mspm{20.69}{3.70} &
\cellcolor{colCFS!51}\mspm{32.37}{1.63} &
\cellcolor{colSSS!78}\mspm{80.87}{2.69} &
\cellcolor{colRFS!49}\mspm{37.79}{2.92} \\

ZTurbo~\cite{cai2025z} &
\cellcolor{colVGQ!74}\mspm{85.10}{1.29} &
\cellcolor{colVEA!77}\mspm{76.42}{4.38} &
\cellcolor{colTGQ!85}\mspm{70.38}{2.07} &
\cellcolor{colTEA!100}\mspm{93.92}{1.79} &
\cellcolor{colLC!21}\mspm{9.00}{2.54} &
\cellcolor{colHGQ!62}\mspm{45.90}{2.06} &
\cellcolor{colPPE!75}\mspm{65.70}{3.50} &
\cellcolor{colPES!91}\mspm{88.52}{2.62} &
\cellcolor{colTAS!65}\mspm{55.40}{4.51} &
\cellcolor{colCFS!21}\mspm{5.87}{1.71} &
\cellcolor{colSSS!96}\mspm{92.96}{0.58} &
\cellcolor{colRFS!54}\mspm{31.04}{3.34} \\

LongCat~\cite{LongCat-Image} &
\cellcolor{colVGQ!66}\mspm{73.75}{1.80} &
\cellcolor{colVEA!87}\mspm{83.97}{2.55} &
\cellcolor{colTGQ!43}\mspm{28.90}{5.11} &
\cellcolor{colTEA!19}\mspm{5.58}{1.60} &
\cellcolor{colLC!83}\mspm{93.30}{1.83} &
\cellcolor{colHGQ!45}\mspm{29.23}{1.71} &
\cellcolor{colPPE!66}\mspm{58.50}{4.35} &
\cellcolor{colPES!61}\mspm{67.88}{3.31} &
\cellcolor{colTAS!21}\mspm{6.08}{1.66} &
\cellcolor{colCFS!56}\mspm{36.89}{3.21} &
\cellcolor{colSSS!86}\mspm{86.89}{1.32} &
\cellcolor{colRFS!39}\mspm{26.74}{2.54} \\

Step1X~\cite{liu2025step1x-edit} &
\cellcolor{colVGQ!62}\mspm{69.95}{2.39} &
\cellcolor{colVEA!73}\mspm{71.20}{3.68} &
\cellcolor{colTGQ!64}\mspm{49.47}{5.06} &
\cellcolor{colTEA!19}\mspm{5.20}{1.53} &
\cellcolor{colLC!99}\mspm{97.90}{1.52} &
\cellcolor{colHGQ!44}\mspm{28.55}{2.74} &
\cellcolor{colPPE!53}\mspm{45.50}{4.81} &
\cellcolor{colPES!51}\mspm{54.65}{3.42} &
\cellcolor{colTAS!19}\mspm{4.41}{1.32} &
\cellcolor{colCFS!66}\mspm{45.22}{3.29} &
\cellcolor{colSSS!77}\mspm{80.50}{2.03} &
\cellcolor{colRFS!37}\mspm{25.13}{3.06} \\

Bagel7B~\cite{deng2025emerging} &
\cellcolor{colVGQ!25}\mspm{32.55}{3.62} &
\cellcolor{colVEA!29}\mspm{23.07}{3.35} &
\cellcolor{colTGQ!20}\mspm{5.10}{2.37} &
\cellcolor{colTEA!15}\mspm{0.58}{0.26} &
\cellcolor{colLC!73}\mspm{82.00}{3.27} &
\cellcolor{colHGQ!19}\mspm{4.55}{1.25} &
\cellcolor{colPPE!30}\mspm{21.20}{3.36} &
\cellcolor{colPES!28}\mspm{23.40}{3.61} &
\cellcolor{colTAS!15}\mspm{0.24}{0.11} &
\cellcolor{colCFS!23}\mspm{7.29}{1.74} &
\cellcolor{colSSS!45}\mspm{50.17}{5.75} &
\cellcolor{colRFS!18}\mspm{4.86}{0.86} \\

GoT~\cite{fang2025got} &
\cellcolor{colVGQ!20}\mspm{26.77}{2.71} &
\cellcolor{colVEA!33}\mspm{28.18}{4.20} &
\cellcolor{colTGQ!49}\mspm{34.88}{3.87} &
\cellcolor{colTEA!16}\mspm{1.53}{0.89} &
\cellcolor{colLC!62}\mspm{65.70}{3.09} &
\cellcolor{colHGQ!21}\mspm{5.88}{1.33} &
\cellcolor{colPPE!32}\mspm{24.10}{4.70} &
\cellcolor{colPES!36}\mspm{33.35}{4.57} &
\cellcolor{colTAS!16}\mspm{0.74}{0.51} &
\cellcolor{colCFS!28}\mspm{11.52}{1.46} &
\cellcolor{colSSS!57}\mspm{62.92}{5.15} &
\cellcolor{colRFS!22}\mspm{8.12}{1.98} \\

IP2P~\cite{brooks2023instructpix2pix} &
\cellcolor{colVGQ!20}\mspm{26.47}{2.64} &
\cellcolor{colVEA!25}\mspm{17.88}{1.90} &
\cellcolor{colTGQ!56}\mspm{42.15}{4.50} &
\cellcolor{colTEA!16}\mspm{1.05}{0.72} &
\cellcolor{colLC!100}\mspm{98.20}{1.87} &
\cellcolor{colHGQ!19}\mspm{4.47}{0.87} &
\cellcolor{colPPE!25}\mspm{13.20}{2.74} &
\cellcolor{colPES!28}\mspm{23.03}{2.94} &
\cellcolor{colTAS!15}\mspm{0.26}{0.21} &
\cellcolor{colCFS!36}\mspm{16.72}{1.81} &
\cellcolor{colSSS!44}\mspm{49.70}{4.58} &
\cellcolor{colRFS!20}\mspm{6.37}{1.49} \\

OmniG2~\cite{wu2025omnigen2} &
\cellcolor{colVGQ!15}\mspm{22.00}{2.25} &
\cellcolor{colVEA!16}\mspm{6.15}{2.91} &
\cellcolor{colTGQ!57}\mspm{42.37}{3.18} &
\cellcolor{colTEA!22}\mspm{9.72}{2.99} &
\cellcolor{colLC!75}\mspm{84.40}{4.60} &
\cellcolor{colHGQ!17}\mspm{2.48}{1.17} &
\cellcolor{colPPE!19}\mspm{9.40}{3.50} &
\cellcolor{colPES!17}\mspm{9.45}{3.22} &
\cellcolor{colTAS!16}\mspm{0.78}{0.55} &
\cellcolor{colCFS!27}\mspm{10.86}{1.48} &
\cellcolor{colSSS!22}\mspm{24.37}{7.13} &
\cellcolor{colRFS!19}\mspm{5.93}{2.13} \\

ICEdit~\cite{zhang2025enabling} &
\cellcolor{colVGQ!22}\mspm{28.82}{3.79} &
\cellcolor{colVEA!15}\mspm{4.80}{0.70} &
\cellcolor{colTGQ!26}\mspm{10.85}{2.95} &
\cellcolor{colTEA!15}\mspm{0.48}{0.48} &
\cellcolor{colLC!80}\mspm{89.70}{2.63} &
\cellcolor{colHGQ!15}\mspm{0.68}{0.44} &
\cellcolor{colPPE!15}\mspm{5.40}{2.37} &
\cellcolor{colPES!15}\mspm{6.17}{1.05} &
\cellcolor{colTAS!15}\mspm{0.04}{0.05} &
\cellcolor{colCFS!20}\mspm{4.53}{2.57} &
\cellcolor{colSSS!15}\mspm{16.87}{2.64} &
\cellcolor{colRFS!15}\mspm{1.94}{0.82} \\

SD~\cite{rombach2021highresolution} &
\cellcolor{colVGQ!85}\mspm{93.92}{1.01} &
\cellcolor{colVEA!88}\mspm{84.88}{3.07} &
\cellcolor{colTGQ!15}\mspm{0.28}{0.25} &
\cellcolor{colTEA!16}\mspm{1.20}{0.59} &
\cellcolor{colLC!15}\mspm{0.40}{0.52} &
\cellcolor{colHGQ!34}\mspm{18.93}{1.71} &
\cellcolor{colPPE!60}\mspm{51.90}{7.19} &
\cellcolor{colPES!93}\mspm{91.00}{1.55} &
\cellcolor{colTAS!16}\mspm{1.22}{0.63} &
\cellcolor{colCFS!15}\mspm{0.03}{0.04} &
\cellcolor{colSSS!97}\mspm{93.47}{0.31} &
\cellcolor{colRFS!16}\mspm{2.32}{0.64} \\

\hline
\end{tabular}%
\begin{tablenotes}[flushleft]
\footnotesize
\item \textit{Notes:}
% Results are reported on the filtered evaluation subset with $\tilde{N}=2711$ memes. For each of the 10 runs, we first randomly sample 100 memes from the full subset and then evaluate all models on this shared sample set. All values are normalized to the range $[0,100]$ and reported as mean $\pm$ standard deviation over the 10 runs. Colors are applied independently within each metric, and darker shading indicates better performance in that column.
Results are reported on the comparable subset with \(\tilde{N}=2{,}711\) memes.
For each of the 10 runs, we randomly sample 100 memes from this subset and evaluate all models on the same sampled instances.
All values are normalized to \([0,100]\) and reported as mean \(\pm\) standard deviation over the 10 runs.
Darker shading indicates better performance within each metric.
\end{tablenotes}
\end{threeparttable}
}
\vspace{-1em}
\end{table*}

\begin{table*}[t]
\setlength{\tabcolsep}{3.0pt}
\renewcommand{\arraystretch}{1.15}
\centering
\caption{Subcategory level performance across evaluation metrics}
\label{tab:category_metric_heat}
\vspace{-0.6em}
\resizebox{\textwidth}{!}{%
\begin{threeparttable}
\begin{tabular}{|l|cccccccc|ccc||c|}
\hline

& \textbf{VGQ} $\uparrow$ & \textbf{VEA} $\uparrow$& \textbf{TGQ} $\uparrow$& \textbf{TEA} $\uparrow$& \textbf{LC} $\uparrow$
& \textbf{HGQ} $\uparrow$& \textbf{PPE} $\uparrow$& \textbf{PES} $\uparrow$
& \textbf{TAS} $\uparrow$& \textbf{CFS} $\uparrow$& \textbf{SSS} $\uparrow$& \textbf{RFS} $\uparrow$\\
\hline

\multicolumn{13}{|l|}{\textbf{\textit{Visual Modality}}} \\

Object/Animal-Centric
& \cellcolor{colVGQ!100}\mspm{65.65}{0.00}
& \cellcolor{colVEA!70}\mspm{59.25}{0.00}
& \cellcolor{colTGQ!100}\mspm{38.75}{0.00}
& \cellcolor{colTEA!72}\mspm{37.20}{0.00}
& \cellcolor{colLC!42}\mspm{74.00}{5.66}
& \cellcolor{colHGQ!88}\mspm{30.25}{0.00}
& \cellcolor{colPPE!61}\mspm{49.60}{8.68}
& \cellcolor{colPES!58}\mspm{61.20}{0.00}
& \cellcolor{colTAS!61}\mspm{22.65}{4.13}
& \cellcolor{colCFS!99}\mspm{31.42}{3.34}
& \cellcolor{colSSS!62}\mspm{83.99}{1.50}
& \cellcolor{colRFS!74}\mspm{38.96}{2.59} \\

Photographic Human
& \cellcolor{colVGQ!59}\mspm{61.25}{0.00}
& \cellcolor{colVEA!78}\mspm{60.05}{0.00}
& \cellcolor{colTGQ!80}\mspm{36.80}{0.00}
& \cellcolor{colTEA!57}\mspm{35.75}{0.00}
& \cellcolor{colLC!100}\mspm{77.80}{2.90}
& \cellcolor{colHGQ!82}\mspm{29.60}{0.00}
& \cellcolor{colPPE!41}\mspm{46.80}{6.68}
& \cellcolor{colPES!62}\mspm{61.65}{0.00}
& \cellcolor{colTAS!44}\mspm{21.03}{3.88}
& \cellcolor{colCFS!100}\mspm{31.48}{2.06}
& \cellcolor{colSSS!66}\mspm{84.23}{1.17}
& \cellcolor{colRFS!65}\mspm{38.14}{2.90} \\

Illustrated/Cartoon
& \cellcolor{colVGQ!64}\mspm{61.75}{0.00}
& \cellcolor{colVEA!27}\mspm{55.30}{0.00}
& \cellcolor{colTGQ!74}\mspm{36.15}{0.00}
& \cellcolor{colTEA!72}\mspm{37.20}{0.00}
& \cellcolor{colLC!94}\mspm{77.40}{2.99}
& \cellcolor{colHGQ!72}\mspm{28.60}{0.00}
& \cellcolor{colPPE!51}\mspm{48.20}{5.20}
& \cellcolor{colPES!28}\mspm{57.25}{0.00}
& \cellcolor{colTAS!49}\mspm{21.48}{3.35}
& \cellcolor{colCFS!93}\mspm{30.90}{2.06}
& \cellcolor{colSSS!29}\mspm{81.93}{2.15}
& \cellcolor{colRFS!62}\mspm{37.86}{2.64} \\

Template Stylized
& \cellcolor{colVGQ!48}\mspm{60.05}{0.00}
& \cellcolor{colVEA!58}\mspm{58.15}{0.00}
& \cellcolor{colTGQ!20}\mspm{30.75}{0.00}
& \cellcolor{colTEA!20}\mspm{32.25}{0.00}
& \cellcolor{colLC!45}\mspm{74.20}{3.71}
& \cellcolor{colHGQ!48}\mspm{26.10}{0.00}
& \cellcolor{colPPE!27}\mspm{45.00}{7.50}
& \cellcolor{colPES!52}\mspm{60.45}{0.00}
& \cellcolor{colTAS!20}\mspm{18.60}{3.18}
& \cellcolor{colCFS!44}\mspm{26.88}{1.26}
& \cellcolor{colSSS!56}\mspm{83.61}{1.75}
& \cellcolor{colRFS!27}\mspm{34.67}{2.28} \\

\multicolumn{13}{|l|}{\textbf{\textit{Structural Layout}}} \\

Single-Panel
& \cellcolor{colVGQ!66}\mspm{61.95}{0.00}
& \cellcolor{colVEA!42}\mspm{56.55}{0.00}
& \cellcolor{colTGQ!72}\mspm{35.95}{0.00}
& \cellcolor{colTEA!35}\mspm{34.65}{0.00}
& \cellcolor{colLC!62}\mspm{75.40}{3.13}
& \cellcolor{colHGQ!74}\mspm{28.15}{0.00}
& \cellcolor{colPPE!41}\mspm{46.80}{5.01}
& \cellcolor{colPES!38}\mspm{58.65}{0.00}
& \cellcolor{colTAS!35}\mspm{20.10}{3.29}
& \cellcolor{colCFS!60}\mspm{29.82}{2.78}
& \cellcolor{colSSS!42}\mspm{82.70}{1.91}
& \cellcolor{colRFS!49}\mspm{36.71}{3.18} \\

Multi-Panel
& \cellcolor{colVGQ!20}\mspm{57.05}{0.00}
& \cellcolor{colVEA!20}\mspm{54.65}{0.00}
& \cellcolor{colTGQ!21}\mspm{30.80}{0.00}
& \cellcolor{colTEA!36}\mspm{33.80}{0.00}
& \cellcolor{colLC!20}\mspm{72.60}{5.74}
& \cellcolor{colHGQ!20}\mspm{23.30}{0.00}
& \cellcolor{colPPE!39}\mspm{46.60}{7.78}
& \cellcolor{colPES!20}\mspm{56.20}{0.00}
& \cellcolor{colTAS!30}\mspm{19.62}{4.79}
& \cellcolor{colCFS!20}\mspm{24.93}{2.43}
& \cellcolor{colSSS!20}\mspm{81.37}{2.11}
& \cellcolor{colRFS!20}\mspm{34.00}{3.65} \\

\multicolumn{13}{|l|}{\textbf{\textit{Affective Polarity}}} \\

Positive Valence
& \cellcolor{colVGQ!89}\mspm{64.43}{0.00}
& \cellcolor{colVEA!100}\mspm{62.07}{0.00}
& \cellcolor{colTGQ!62}\mspm{34.93}{0.00}
& \cellcolor{colTEA!100}\mspm{39.89}{0.00}
& \cellcolor{colLC!74}\mspm{74.94}{6.51}
& \cellcolor{colHGQ!100}\mspm{31.42}{0.00}
& \cellcolor{colPPE!100}\mspm{54.92}{5.44}
& \cellcolor{colPES!100}\mspm{66.68}{0.00}
& \cellcolor{colTAS!100}\mspm{26.58}{3.20}
& \cellcolor{colCFS!93}\mspm{30.92}{3.92}
& \cellcolor{colSSS!100}\mspm{86.38}{1.61}
& \cellcolor{colRFS!100}\mspm{41.35}{2.86} \\

Neutral Valence
& \cellcolor{colVGQ!48}\mspm{60.10}{0.00}
& \cellcolor{colVEA!34}\mspm{55.95}{0.00}
& \cellcolor{colTGQ!41}\mspm{32.80}{0.00}
& \cellcolor{colTEA!42}\mspm{34.30}{0.00}
& \cellcolor{colLC!20}\mspm{72.60}{4.90}
& \cellcolor{colHGQ!65}\mspm{26.90}{0.00}
& \cellcolor{colPPE!20}\mspm{44.00}{7.18}
& \cellcolor{colPES!55}\mspm{59.40}{0.00}
& \cellcolor{colTAS!22}\mspm{18.83}{4.30}
& \cellcolor{colCFS!47}\mspm{27.23}{3.35}
& \cellcolor{colSSS!79}\mspm{82.89}{3.38}
& \cellcolor{colRFS!28}\mspm{34.73}{3.45} \\

Negative Valence
& \cellcolor{colVGQ!73}\mspm{62.80}{0.00}
& \cellcolor{colVEA!74}\mspm{59.60}{0.00}
& \cellcolor{colTGQ!58}\mspm{34.60}{0.00}
& \cellcolor{colTEA!31}\mspm{33.25}{0.00}
& \cellcolor{colLC!55}\mspm{75.80}{4.26}
& \cellcolor{colHGQ!72}\mspm{27.10}{0.00}
& \cellcolor{colPPE!64}\mspm{49.40}{9.71}
& \cellcolor{colPES!49}\mspm{59.85}{0.00}
& \cellcolor{colTAS!45}\mspm{21.16}{5.21}
& \cellcolor{colCFS!75}\mspm{29.47}{3.44}
& \cellcolor{colSSS!52}\mspm{83.25}{2.28}
& \cellcolor{colRFS!46}\mspm{37.16}{4.22} \\

\hline
\end{tabular}%
\begin{tablenotes}[flushleft]
\footnotesize
\item \textit{Notes:}
Results report category-wise performance on the comparable subset with \(\tilde{N}=2{,}711\) memes.
For each subcategory, we randomly sample up to 50 memes from the comparable subset and evaluate all models on the same sampled instances.
This process is repeated 10 times, and results are reported as mean \(\pm\) standard deviation over the 10 runs.
All values are normalized to \([0,100]\).
Darker shading indicates better performance within each metric.
% Results report aggregated performance for each dataset subcategory. For each category, we sample 50 memes as evenly as possible from the edited outputs produced by different models, ensuring balanced coverage across models. This sampling procedure is repeated 10 times. All values are normalized to the range $[0,100]$ and reported as mean $\pm$ standard deviation across the 10 runs. Colors are applied independently within each metric, and darker shading indicates better performance within that column.
\end{tablenotes}
\end{threeparttable}
}
\vspace{-1em}
\end{table*}

\subsection{Aggregated Evaluation Score}
\label{subsec:aggregated-score}
Although the eight metrics provide fine-grained diagnostics, MER requires three objectives to hold jointly: target emotional alignment, content fidelity, and perceptible affective transformation.
A model should not obtain a high overall score by excelling in only one dimension, such as visual quality, while failing to preserve structure or reach the target emotion.
We therefore aggregate the metrics into three intermediate scores and a final conjunctive score.
All scalar metrics are normalized to $[0,1]$ before aggregation.
Let $H\in\{0,1\}$ be the target emotion hit indicator, where $H=1$ if the perceived primary emotion matches the target emotion and $H=0$ otherwise.
Let $E_v$ and $E_t$ denote normalized visual and textual emotion alignment scores, $Q_v$, $Q_t$, and $Q_o$ denote normalized visual, textual, and holistic generation quality scores, $C$ denote normalized layout consistency, and $S$ denote normalized perceived emotion shift.

\paragraph{Target Alignment Score (TAS).}
TAS measures whether the edited meme realizes the target emotion across visual and textual modalities:
\begin{equation}
\mathrm{TAS}=H\cdot \mathrm{HM}(E_v,E_t),
\end{equation}
where $\mathrm{HM}(x,y)=\frac{2xy}{x+y+\epsilon}$ is the harmonic mean.
The target-hit gate suppresses the score when the perceived emotion misses the target, while the harmonic mean penalizes modality imbalance.

\paragraph{Content Fidelity Score (CFS).}
CFS measures semantic, structural, and generation-quality preservation:
\begin{equation}
\mathrm{CFS}=C\cdot \mathrm{GM}(Q_v,Q_t,Q_o),
\end{equation}
where $\mathrm{GM}(\cdot)$ denotes the geometric mean.
The multiplicative layout term directly penalizes structural inconsistency.

\paragraph{Shift Strength Score (SSS).}
SSS quantifies the magnitude of target-consistent emotional transformation:
\begin{equation}
\mathrm{SSS}=\frac{1-\exp(-\alpha S)}{1-\exp(-\alpha)},
\end{equation}
where $\alpha>0$ controls the saturation rate.
This normalized form maps $S=0$ to 0 and $S=1$ to 1.

\paragraph{Reappraisal Fidelity Score (RFS).}
Finally, we combine the three components using an $\epsilon$-smoothed geometric conjunction:
\begin{equation}
\mathrm{RFS}
=
\left[
(\mathrm{TAS}+\epsilon)
(\mathrm{CFS}+\epsilon)
(\mathrm{SSS}+\epsilon)
\right]^{1/3}
-\epsilon.
\end{equation}

This soft conjunctive operator preserves the requirement that all three objectives should be satisfied, while avoiding numerical instability when one component is near zero.
Thus, RFS reflects the logical structure of MER: strong reappraisal requires simultaneous target alignment, content fidelity, and affective transformation.

\subsection{Evaluation Procedure}
\label{subsec:model-benchmarking}
We evaluate 14 image-editing and multimodal-generation models spanning diffusion-based editors, instruction-guided frameworks, MLLM-conditioned generators, and large-scale image transformation systems.
For each source meme and target emotion, a model generates one edited meme, which is independently evaluated using the MLLM-as-a-Judge protocol in Sec.~\ref{subsec:mllm-as-a-judge}.
In total, \(M=14\) models are evaluated on \(N=3{,}117\) MER-Bench samples.
Some models fail on certain inputs due to missing images, corrupted files, unreadable or missing meme text, safety refusals, or severe format violations.
We treat only such technical or format failures as invalid; low-quality but scorable outputs are retained and penalized by the metrics.
For fair comparison, we retain the intersection subset where all models produce valid outputs, yielding \(\tilde{N}=2{,}711\) comparable memes.
We adopt a repeated randomized subset evaluation protocol, with scores computed at the output level and aggregated by model, taxonomy category, and overall performance.
Unless otherwise specified, all main tables use this comparable subset to avoid sample-composition shifts across models.

\section{Quantitative Results}
\label{sec:quatitative-results}
\begin{figure*}[t!]
\centering
\includegraphics[width=0.8\textwidth,height=\textheight,keepaspectratio]{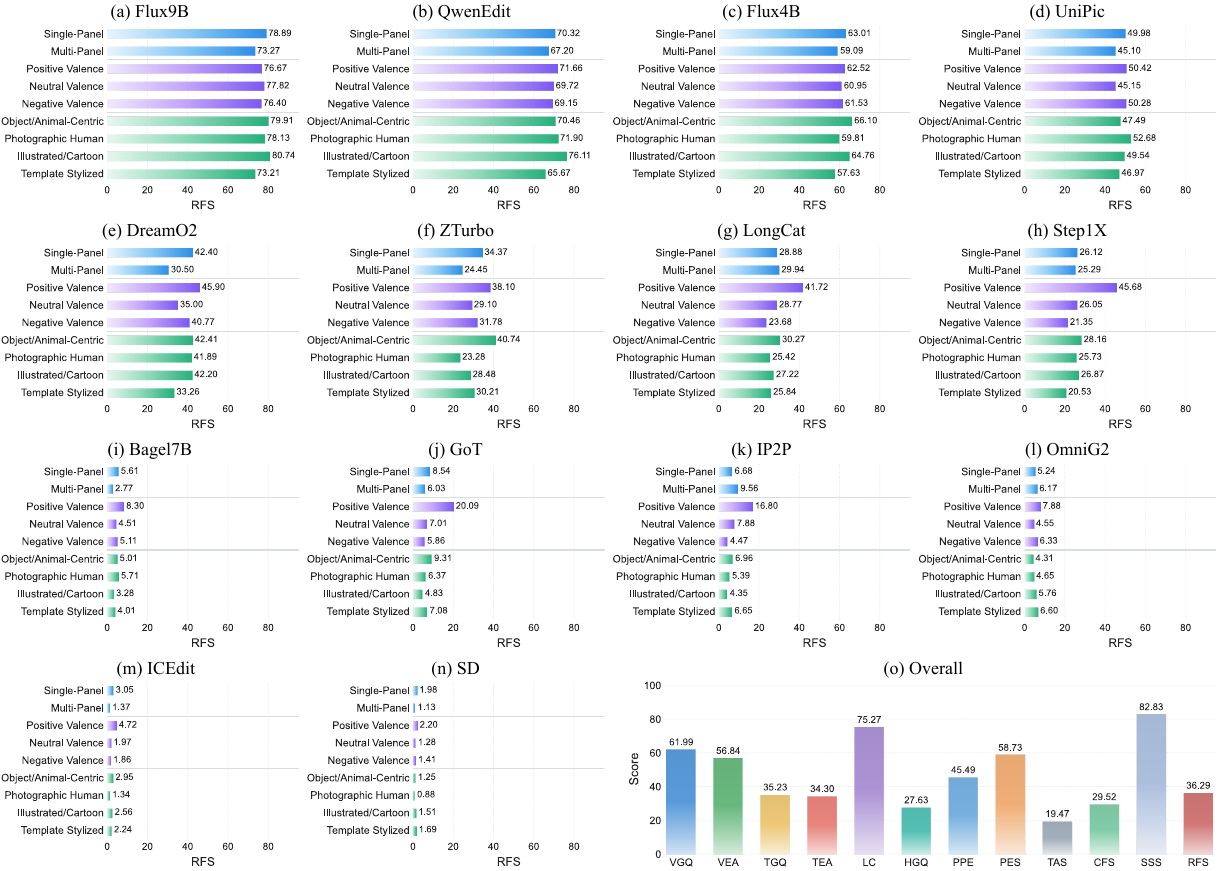}
% \vspace{-2.0em}
\caption{
Subcategory-wise and overall analysis on Multimodal Emotional Reappraisal. 
In each run, 700 memes are sampled as evenly as possible from the edited outputs of different models, and scores are averaged over 10 runs.
The bar colors in (o) are consistent with the metric-specific color scheme used in Tables~\ref{tab:main_results_sorted_heat} and~\ref{tab:category_metric_heat}.
% Subcategory-wise and overall analysis on Multimodal Emotional Reappraisal. For each run, 100 memes are randomly sampled from the comparable subset and evaluated across all models; scores are averaged over 10 runs. The bar colors in (o) follow the metric-specific color scheme used in Tables~\ref{tab:main_results_sorted_heat} and~\ref{tab:category_metric_heat}.
}
\vspace{-1em}
\label{fig:model-cls}
\end{figure*}
\subsection{Model-wise Performance Analysis}
Table~\ref{tab:main_results_sorted_heat} reports metric-level results for 14 models detailed in Table~\ref{tab:selected-models}, while Fig.~\ref{fig:model-cls}(a)-(n) shows subcategory-wise RFS scores.
\model{Flux9B}~\cite{flux-2-2025} achieves the best overall RFS of 76.78, followed by \model{QwenEdit}~\cite{wu2025qwenimagetechnicalreport} with 69.38 and \model{Flux4B}~\cite{flux-2-2025} with 61.06.
These top models maintain relatively balanced performance across emotional alignment, structural preservation, and generation quality.
Notably, although \model{SD} attains the highest VGQ, its near-zero text and layout scores result in a very low RFS, indicating that visual quality alone is insufficient for MER.
The middle tier, including \model{UniPic}~\cite{wei2026skywork}, \model{DreamO2}~\cite{xia2025dreamomni2}, and \model{ZTurbo}~\cite{cai2025z}, exhibits strong but uneven capabilities.
For example, \model{ZTurbo} achieves high text quality and emotional alignment, but its extremely low LC indicates severe structural disruption.
Conversely, instruction-based editors such as \model{Step1X}~\cite{liu2025step1x-edit} and \model{IP2P}~\cite{brooks2023instructpix2pix} preserve layout well but show limited emotional transformation.
Traditional diffusion baselines and smaller MLLMs, including \model{Bagel7B}~\cite{deng2025emerging}, \model{GoT}~\cite{fang2025got}, \model{OmniG2}~\cite{wu2025omnigen2}, and \model{ICEdit}~\cite{zhang2025enabling}, generally fail to satisfy the joint requirements of MER, especially in text generation and affective alignment.

\textbf{Key Insights.}
Model-wise results show that MER is not bottlenecked by low-level visual generation, but by the ability to coordinate affective transformation with semantic and structural preservation.
Current systems often optimize only one side of this trade-off: \model{ZTurbo} improves affective and textual cues at the cost of layout consistency, while \model{Step1X} and \model{IP2P} preserve templates without sufficiently changing affective framing.
The strong performance of \model{Flux9B} and \model{QwenEdit} suggests that effective Emotional Reappraisal requires integrated multimodal reasoning rather than isolated image editing or text rewriting.

\subsection{Category-wise Performance Analysis}

Tab.~\ref{tab:category_metric_heat} summarizes performance across dataset subcategories, and Fig.~\ref{fig:model-cls}(a)-(n) shows the corresponding RFS distributions.
The results reveal that MER difficulty varies substantially with visual modality, layout structure, and affective polarity.
Across visual modalities, object/animal-centric memes achieve the highest RFS of 38.96, followed by photographic human (38.14) and illustrated/cartoon memes (37.86), while template-stylized memes perform worst (34.67).
This drop is mainly reflected in TGQ and TEA, suggesting that fixed templates and stylized captions make text rewriting more error-prone and semantically fragile.
For structural layout, single-panel memes outperform multi-panel memes (36.71 \textit{vs.} 34.00), as multi-panel memes require preserving both spatial organization and cross-panel narrative coherence.
Affective polarity also affects performance: positive-valence memes achieve the highest RFS (41.35), followed by negative (37.16) and neutral memes (34.73), indicating that ambiguous or weakly expressed affect makes controlled reappraisal harder.

\textbf{Key Insights.}
Category-wise results show that MER becomes difficult when meme meaning is tied to rigid templates, sequential layouts, or ambiguous affective cues.
These findings suggest that the main challenge is not generating positive content, but preserving the structural and pragmatic conditions under which the original meme conveys meaning.

\begin{figure}[!t]
\centering
% width=\columnwidth 确保图片自动适配单栏的宽度
\includegraphics[width=\columnwidth]{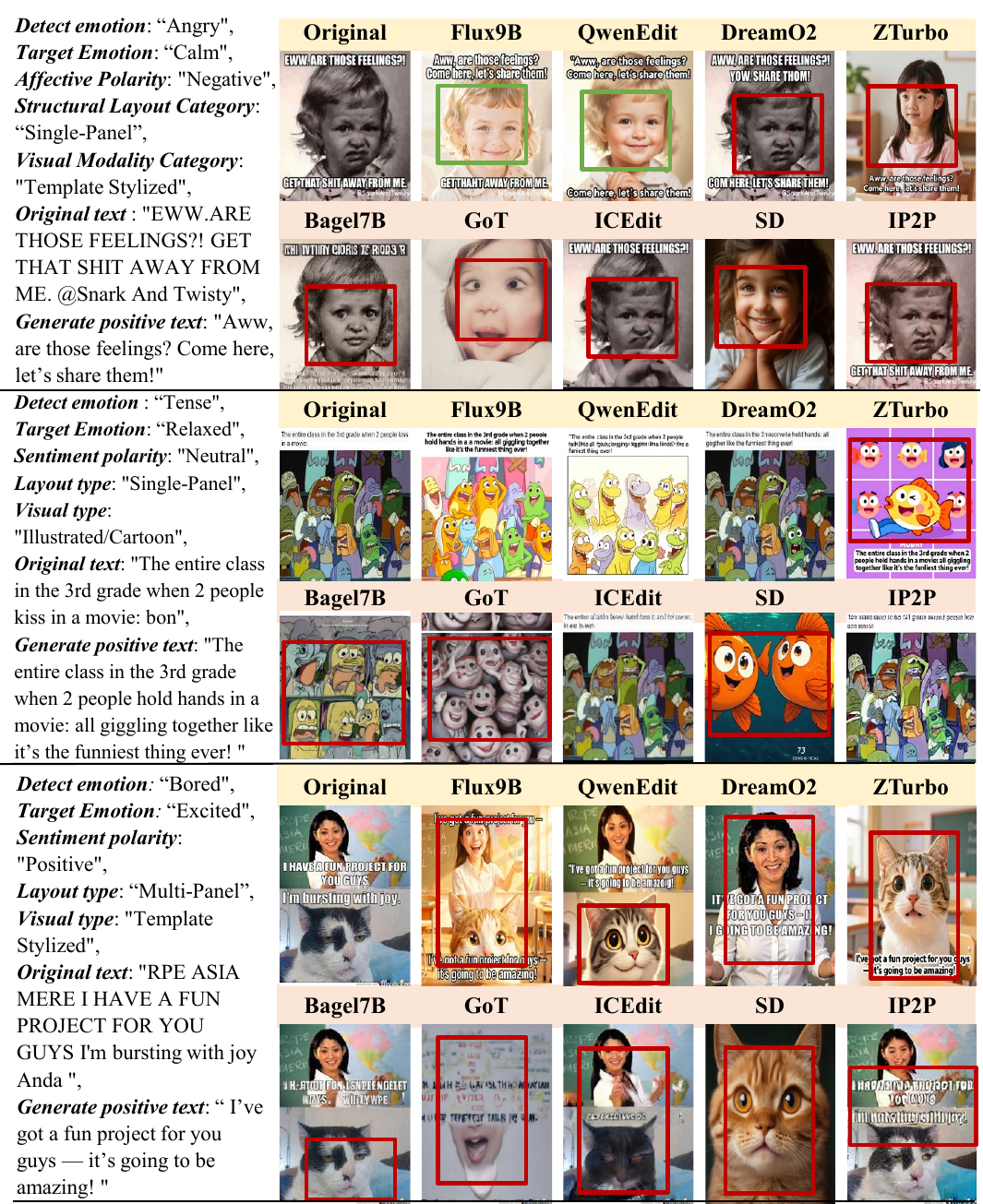}
\vspace{-1.8em}
\caption{Qualitative comparison of MER results across models on MER-Bench.}
\label{fig:qualitative_meme_reappraisal}
\vspace{-1.5em}
\end{figure}

\subsection{Overall Benchmark Analysis}
Fig.~\ref{fig:model-cls}(o) summarizes the overall metric distribution across the filtered evaluation subset.
The global results reveal a clear gap between low-level generation quality and higher-level reappraisal capability.
Visual quality and basic structural metrics, such as VGQ and LC, remain relatively strong across models, suggesting that current systems can often produce visually coherent outputs and preserve coarse layouts.
In contrast, text- and affect-related metrics, including TGQ, TEA, and TAS, are substantially weaker, indicating persistent difficulty in rewriting embedded text and aligning the edited meme with the target affect.
Notably, the high SSS but low TAS, CFS, and RFS suggest that models can often induce visible emotional changes, but struggle to make them target-aligned and content-faithful.
% Notably, SSS is much higher than TAS and CFS, suggesting that models can often induce perceptible emotional changes, but these shifts are not reliably aligned with the target emotion or grounded in faithful content preservation.
% The resulting low RFS further confirms that successful MER requires the conjunction of affective shift, target alignment, and content fidelity rather than improvement along a single dimension.

\textbf{Key Insights.}
Overall results show that MER is not primarily limited by image synthesis quality, but by the ability to coordinate textual rewriting, affective transformation, and semantic preservation under structural constraints.
This confirms that MER is a multimodal reasoning problem rather than a purely visual editing task.

\section{Qualitative Analysis}
\label{sec:qualitative-results}
Fig.~\ref{fig:qualitative_meme_reappraisal} shows examples of common failure modes that are consistent with quantitative trends.
Models often preserve the template and layout but fail to induce an affective shift toward the target emotion.
This pattern appears in \model{Bagel7B}~\cite{deng2025emerging}, \model{GoT}~\cite{fang2025got}, \model{SD}~\cite{rombach2021highresolution}, and \model{IP2P}~\cite{brooks2023instructpix2pix}, where outputs remain close to the source meme yet weakly align with the target emotion.
A contrasting failure occurs when models achieve the desired affective shift at the cost of structural preservation.
For example, \model{ZTurbo}~\cite{cai2025z} often produces emotionally plausible edits but disrupts the meme template, caption placement, or layout.
Multi-panel memes are harder, as they require preserving spatial organization, panel order, and cross-panel narrative coherence.
Many models distort inter-panel relations or produce inconsistent content, leading to semantic drift despite plausible visuals.

\textbf{Key Insights.}
The qualitative results reveal that current models often fail at the interface between structure, text, and affect.
Preserving layouts alone does not guarantee emotional transformation, while affective edits frequently introduce caption rendering errors, layout distortion, or semantic drift.
This confirms that MER is not merely an image-editing problem but a coordinated multimodal reasoning challenge involving visual structure, embedded text, semantic intent, and emotional framing.

\section{Conclusion}
We study \textbf{Multimodal Emotional Reappraisal (MER)} as a structure-preserving affective transformation problem for image-text memes, and introduce \textbf{\textit{MER-Bench}} for evaluation.
MER-Bench provides annotations and a human-validated MLLM-as-a-Judge evaluation suite.
Experiments on 14 multimodal generation systems show that models still struggle to jointly achieve emotional transformation, semantic preservation, and structural fidelity.
These findings position MER as a challenging multimodal reasoning problem beyond conventional image editing or sentiment transfer, and we hope MER-Bench will support future research on controllable affect-aware generation.

\section*{Limitations}
\label{sec:limitations}

MER-Bench focuses on memes, leaving dynamic formats such as GIFs, short videos, and audio-visual memes unexplored.
It also does not explicitly address multilingual or cross-cultural variation, despite the influence of language, culture, and community conventions on meme interpretation and emotional reappraisal.
Finally, although expert-validated, our MLLM-as-a-Judge protocol may still inherit biases from the underlying judge model.
Future work will extend MER toward multilingual, multicultural, and temporally grounded settings.

\section*{Ethical Considerations}

MER-Bench is built from available meme data for research on controllable, affect-aware multimodal generation.
It focuses on transforming negative affect into more positive or constructive interpretations rather than generating harmful, deceptive, or manipulative content.
Unsafe or infeasible samples are filtered, and retained annotations are manually verified to reduce misuse.
Although downstream risks cannot be eliminated, MER-Bench is designed to promote safer affective transformation.

\section*{Acknowledgment}
This work was supported by the Yangtze River Delta Science and Technology Innovation Community Joint Research (Basic Research) Project (2025CSJZN01600) and the New Generation of Artificial Intelligence National Science and Technology Major Project (2025ZD0123303). 

% \clearpage
\bibliography{custom}

\clearpage

\appendix
\section*{Supplementary Material}
\addtocontents{toc}{\protect\setcounter{tocdepth}{2}}
\tableofcontents
\renewcommand{\thetable}{S\arabic{table}}
\renewcommand{\thefigure}{S\arabic{figure}}

\begin{figure*}[t]
\centering
\includegraphics[width=\textwidth]{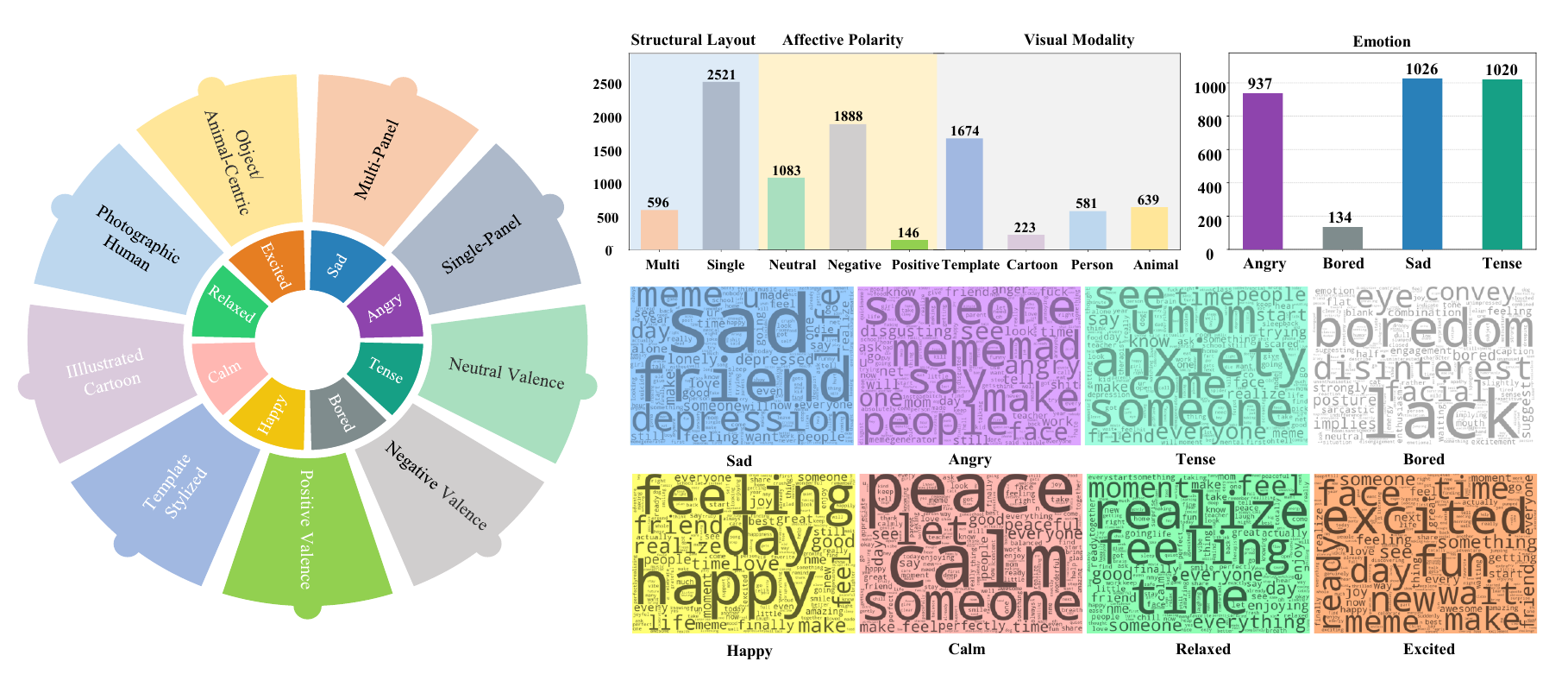}
\caption{\textbf{Dataset statistics of MER-Bench.}
The figure summarizes the distribution of structural layouts, caption-level affective polarities, visual modalities, and source emotion categories.
It also visualizes frequent terms in source and reappraised meme captions through emotion-specific word clouds.
}
\label{fig:data_appendix}
\end{figure*}

\section{Dataset Construction Details}
\label{app:dataset-details}

This appendix provides additional details on MER-Bench construction, including filtering statistics, annotation schema, LLM-assisted human control, inter-annotator agreement, and emotion-shift distribution.

\subsection{Filtering Statistics}

MER-Bench is constructed from the \textit{MOOD} meme corpus~\cite{sharma2024emotion}.
As shown in Table~\ref{tab:filtering-statistics}, we start from 10,005 original memes and retain 6,695 samples with clearly identifiable negative affective framing.
After quality filtering, deduplication, safety filtering, and reappraisal-feasibility filtering, the final benchmark contains 3,117 image-text memes.

\begin{table*}[t]
\centering
\small
\tabcolsep 8pt
\resizebox{1.0\linewidth}{!}{
\begin{tabular}{|l|r|r|l|}
\hline
Stage & Remaining & Removed & Main Criterion \\
\hline
Original MOOD corpus & 10,005 & - & Original meme corpus \\
Negative-affect filtering & 6,695 & 3,310 & Remove non-negative or affectively unclear samples \\
Quality filtering & 5,677 & 1,018 & Remove low-resolution, corrupted, or unreadable memes \\
Deduplication & 4,589 & 1,088 & Remove duplicate or near-duplicate samples \\
Safety filtering & 3,854 & 735 & Remove sensitive or inappropriate content \\
Reappraisal-feasibility filtering & 3,117 & 737 & Remove samples unsuitable for reliable reappraisal \\
\hline
\end{tabular}}
\caption{Filtering statistics for MER-Bench construction.}
\label{tab:filtering-statistics}
\end{table*}

\subsection{Annotation Schema}

Each retained sample contains structured annotations for source understanding, target reappraisal, and multimodal editing.
Table~\ref{tab:annotation-schema} summarizes the annotation schema.

\begin{table*}[t]
\centering
\tabcolsep 8pt
\resizebox{1.0\linewidth}{!}{
\begin{tabular}{|l|l|l|}
\hline
Field & Description & Purpose \\
\hline
id 
& Original meme identifier inherited from MOOD 
& Trace the source sample \\

img\_desc 
& Brief image description covering scene, entities, expressions, and poses 
& Support visual understanding and editing \\

txt 
& Original overlaid meme text 
& Preserve the source textual content \\

src\_emo 
& Source emotion label 
& Identify the original affective framing \\

tgt\_emo 
& Target emotion label 
& Define the intended reappraisal direction \\

rewritten\_caption 
& Reappraised meme text 
& Provide the target textual framing \\

edit\_spec 
& Visual editing specification 
& Guide structure-preserving multimodal editing \\

gen\_id 
& Generated meme identifier 
& Track the generated output \\

notes 
& Optional annotation or quality-control notes 
& Record ambiguity, revision, or failure cases \\
\hline
\end{tabular}}
\caption{Annotation schema of MER-Bench.}
\label{tab:annotation-schema}
\end{table*}
% \begin{table*}[t]
% \centering
% \small
% \begin{tabular}{lll}
% \hline
% Field & Description & Purpose \\
% \hline
% \texttt{id} 
% & Original meme identifier inherited from MOOD 
% & Trace the source sample \\

% \texttt{img\_desc} 
% & Brief image description covering scene, entities, expressions, and poses 
% & Support visual understanding and editing \\

% \texttt{txt} 
% & Original overlaid meme text 
% & Preserve the source textual content \\

% \texttt{src\_emo} 
% & Source emotion label 
% & Identify the original affective framing \\

% \texttt{tgt\_emo} 
% & Target emotion label 
% & Define the intended reappraisal direction \\

% \texttt{rewritten\_caption} 
% & Reappraised meme text 
% & Provide the target textual framing \\

% \texttt{edit\_spec} 
% & Visual editing specification 
% & Guide structure-preserving multimodal editing \\

% \texttt{gen\_id} 
% & Generated meme identifier 
% & Track the generated output \\

% \texttt{notes} 
% & Optional annotation or quality-control notes 
% & Record ambiguity, revision, or failure cases \\
% \hline
% \end{tabular}
% \caption{Annotation schema of MER-Bench.}
% \label{tab:annotation-schema}
% \end{table*}

\subsection{Prompt Templates}
\label{app:prompt-templates}

We use three prompt templates for LLM-assisted human rewriting. 
The templates are normalized here using the annotation field names in Table~\ref{tab:annotation-schema}.
The first two prompts are used with \texttt{Gemini-2.5-Flash}, and the third prompt is used with \texttt{Gemini-2.5-Flash-Image}.
All outputs are manually checked before inclusion in MER-Bench.

\paragraph{Prompt I: Emotion Detection and Image Description.}

\begin{quote}
\small
For the given meme, consisting of an image and overlaid text, complete the following tasks.

\textbf{Emotion labeling.}
Choose one source emotion label from:
\texttt{sad}, \texttt{angry}, \texttt{tense}, or \texttt{bored}.

Use the following emotion definitions:
\texttt{sad} = loss or disappointment;
\texttt{angry} = frustration or unfairness;
\texttt{tense} = pressure or worry;
\texttt{bored} = lack of interest.

When multiple people appear, focus on the main subject whose expression, posture, or role best conveys the meme's emotion.
If the visual expression and the overlaid text conflict, follow the emotion implied by the text.
Always output only one source emotion.

\textbf{Image description.}
Briefly describe the meme using the following information:
visual cues, including facial expression, gesture, color, scene, and layout;
textual cues, including tone and phrasing;
and how these cues support the selected source emotion.

\textbf{Output format.}
Return a JSON object with the following fields:
\texttt{src\_emo} and \texttt{img\_desc}.
\end{quote}

\paragraph{Prompt II: Reappraisal Planning.}

\begin{quote}
\small
Rewrite the meme toward its target reappraisal emotion while preserving the original topic, entities, and core situation.

\textbf{Inputs.}
\texttt{src\_emo}: one of \texttt{sad}, \texttt{angry}, \texttt{tense}, or \texttt{bored};  
\texttt{img\_desc}: a concise description of the meme's visual and textual cues;  
\texttt{txt}: the original overlaid meme text.

\textbf{Emotion mapping.}
\texttt{sad} $\rightarrow$ \texttt{happy};  
\texttt{angry} $\rightarrow$ \texttt{calm};  
\texttt{tense} $\rightarrow$ \texttt{relaxed};  
\texttt{bored} $\rightarrow$ \texttt{excited}.

\textbf{Tasks.}
First, set \texttt{tgt\_emo} to the mapped target emotion.
Second, write \texttt{rewritten\_caption}, a reappraised meme caption that expresses the target emotion while keeping the same topic and context.
Do not change the core situation, entities, or event relations.
Avoid adding hostility, anxiety, fear, sexual content, violent content, or unrelated negativity.
Third, write \texttt{edit\_spec}, a concise visual editing specification describing changes in facial expression, posture, gesture, atmosphere, and visual tone.

\textbf{Tone guide.}
\texttt{happy}: warm, optimistic, joyful;  
\texttt{calm}: peaceful, balanced, stable;  
\texttt{relaxed}: easy-going, comfortable, natural;  
\texttt{excited}: energetic, playful, enthusiastic.

\textbf{Output format.}
Return a JSON object with the following fields:
\texttt{tgt\_emo}, \texttt{rewritten\_caption}, and \texttt{edit\_spec}.
\end{quote}

\paragraph{Prompt III: Reappraised Meme Generation.}

\begin{quote}
\small
Generate a final reappraised meme image and return it as a base64-encoded PNG.

\textbf{Inputs.}
\texttt{source\_image}: the original meme image;  
\texttt{rewritten\_caption}: the target meme text;  
\texttt{edit\_spec}: the visual editing specification;  
\texttt{tgt\_emo}: one of \texttt{happy}, \texttt{calm}, \texttt{relaxed}, or \texttt{excited}.

\textbf{Generation rules.}
Preserve the original meme's core scenario, main characters, and essential layout structure.
Apply the visual changes described in \texttt{edit\_spec} so that the generated image matches \texttt{tgt\_emo}.
Overlay \texttt{rewritten\_caption} directly on the image as meme text, using bold white text with a dark outline.
Keep the output in English.
Do not introduce hostility, aggression, fear, sexual content, violent content, or unrelated objects.

\textbf{Emotion guide.}
\texttt{happy}: warm, joyful, smiling, open gestures;  
\texttt{calm}: peaceful, balanced, gentle smile;  
\texttt{relaxed}: comfortable, natural, easy-going;  
\texttt{excited}: energetic, playful, enthusiastic.

\textbf{Output format.}
Return only the generated meme image as a base64 PNG.
The generated file is stored using the field \texttt{gen\_id}.
\end{quote}

\subsection{Human Control over LLM Assistance}

LLMs are used as annotation assistants rather than automatic data generators.
Human annotators verify, revise, or reject LLM-produced intermediate outputs before a sample is included in MER-Bench.
Table~\ref{tab:human-control} summarizes the human control statistics over the 6,695 negative-affect candidates.

\begin{table}[t]
\centering
\tabcolsep 8pt
\resizebox{1.0\linewidth}{!}{
\begin{tabular}{|l|r|r|}
\hline
Outcome & \# Samples & Ratio \\
\hline
Accepted without modification & 1,126 & 16.8\% \\
Accepted after minor revision & 1,421 & 21.2\% \\
Accepted after substantial rewriting & 570 & 8.5\% \\
Removed after human verification & 3,578 & 53.4\% \\
\hline
Total checked & 6,695 & 100.0\% \\
\hline
\end{tabular}}
\caption{Human control statistics for LLM-assisted annotation.}
\label{tab:human-control}
\end{table}

The retained samples consist of those accepted without modification, accepted after minor revision, and accepted after substantial rewriting, resulting in 3,117 final samples.
The removed samples correspond to candidates filtered out due to quality, duplication, safety, or reappraisal-feasibility issues.

\subsection{Inter-annotator Agreement}

To assess annotation reliability, we compute inter-annotator agreement using Fleiss' $\kappa$ over the four annotators assigned to each sample.
Agreement is computed separately for emotion labeling, taxonomy labeling, and quality-control decisions.
As shown in Table~\ref{tab:annotation-agreement}, the agreement scores range from 0.69 to 0.81, indicating substantial consistency among annotators.

\begin{table}[t]
\centering
\tabcolsep 24pt
\resizebox{1.0\linewidth}{!}{
\begin{tabular}{|l|c|}
\hline
Annotation Task & Fleiss' $\kappa$ \\
\hline
Source emotion labeling & 0.78 \\
Target emotion verification & 0.81 \\
Visual modality labeling & 0.74 \\
Layout structure labeling & 0.76 \\
Affective polarity labeling & 0.72 \\
Quality-control decision & 0.69 \\
\hline
Average & 0.75 \\
\hline
\end{tabular}}
\caption{Inter-annotator agreement for MER-Bench annotations.}
\label{tab:annotation-agreement}
\end{table}

\begin{figure*}[t]
\centering
\includegraphics[width=\textwidth]{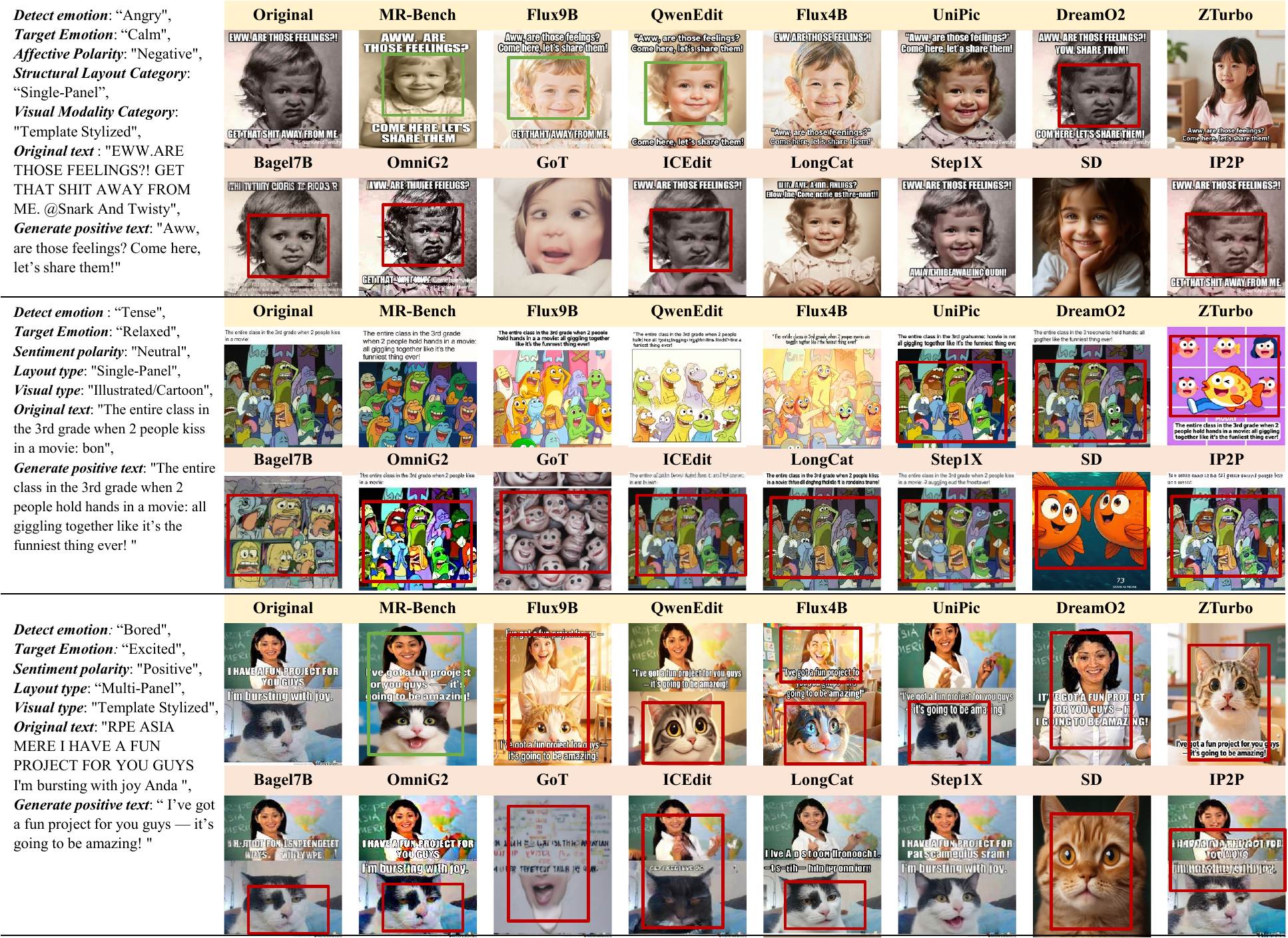}
\caption{\textbf{Full qualitative comparison across evaluated models.}
This figure extends the qualitative examples in Fig.~\ref{fig:qualitative_meme_reappraisal} by showing outputs from all evaluated models for representative source memes.
The examples illustrate typical MER behaviors, including successful reappraisal, insufficient affective shift, layout disruption, text-rendering errors, and semantic drift.
}
\label{fig:appendex_usecase}
\end{figure*}

\subsection{Dataset Statistics}
\label{app:dataset-statistics}

Fig.~\ref{fig:data_appendix} provides additional statistics of MER-Bench, including structural layout, caption-level affective polarity, visual modality, source emotion distribution, and word clouds for source and reappraised captions.
The benchmark covers diverse meme formats, with single-panel memes dominating the structural distribution and template-stylized memes forming the largest visual modality category.
The source emotion distribution corresponds to the four reappraisal directions used in MER-Bench:
\textit{Sad} $\rightarrow$ \textit{Happy},
\textit{Angry} $\rightarrow$ \textit{Calm},
\textit{Tense} $\rightarrow$ \textit{Relaxed}, and
\textit{Bored} $\rightarrow$ \textit{Excited}.
The \textit{Bored} $\rightarrow$ \textit{Excited} category is smaller because bored memes with clear image-text grounding and feasible reappraisal targets are less frequent after filtering.

\section{Model Details}
\label{app:model-details}
Table~\ref{tab:selected-models} lists the 14 models evaluated in MER-Bench, including their full model names, short names, model categories, and editing paradigms.
\begin{table*}[t]
\centering
\caption{Details of models evaluated on MER-Bench.}
\label{tab:selected-models}
\resizebox{\textwidth}{!}{%
\begin{threeparttable}
\begin{tabular}{|l|l|l|l|}
\hline
Model Name (Short Name) & Size & Category & Editing Paradigm \\
\hline
black-forest-labs/FLUX.2-klein-9B (\textbf{Flux9B})~\cite{flux-2-2025}           
& 9B  & Diffusion editing model & Instruction-based image editing \\
Qwen/Qwen-Image-Edit-2509 (\textbf{QwenEdit})~\cite{wu2025qwenimagetechnicalreport}    
& --  & Vision-language editing model & Multimodal instruction editing  \\
black-forest-labs/FLUX.2-klein-4B (\textbf{Flux4B})~\cite{flux-2-2025}           
& 4B  & Diffusion editing model & Instruction-based image editing \\
Skywork/Unipic3 (\textbf{UniPic})~\cite{wei2026skywork}                 
& 1.5B & Vision-language editing model & Multimodal instruction editing  \\
xiabs/DreamOmni2 (\textbf{DreamO2})~\cite{xia2025dreamomni2}             
& -- & Vision-language editing model & Multimodal instruction editing  \\
Tongyi-MAI/Z-Image-Turbo (\textbf{ZTurbo})~\cite{cai2025z}          
& 6B  & Vision-language editing model & Multimodal image editing \\
meituan-longcat/LongCat-Image (\textbf{LongCat})~\cite{LongCat-Image}             
& 6B  & Vision-language editing model & Multimodal image editing \\
stepfun-ai/Step1X-Edit-v1p2 (\textbf{Step1X})~\cite{liu2025step1x-edit}      
& -- & Diffusion editing model & Instruction-based image editing \\
ByteDance-Seed/BAGEL-7B-MoT (\textbf{Bagel7B})~\cite{deng2025emerging}             
& 7B  & Vision-language editing model & Multimodal instruction editing \\
GoT (\textbf{GoT})~\cite{fang2025got}                            
& -- & Vision-language editing model & Multimodal instruction editing \\
timbrooks/instruct-pix2pix (\textbf{IP2P})~\cite{brooks2023instructpix2pix}             
& -- & Diffusion editing model & Text-guided image-to-image \\
OmniGen2/OmniGen2 (\textbf{OmniG2})~\cite{wu2025omnigen2}                 
& -- & Diffusion editing model & Text-guided image-to-image \\
In-Context Edit (\textbf{ICEdit})~\cite{zhang2025enabling}          
& -- & In-context editing baseline & Prompt-only editing \\
stable\_diffusion (\textbf{SD})~\cite{rombach2021highresolution}              
& -- & Diffusion generation model & Text-guided image-to-image \\
\hline
\end{tabular}

\begin{tablenotes}[flushleft]
\footnotesize
\item \textit{Notes:} The table lists the full model names used in our benchmark and the short names used throughout the paper. 
The size column reports the approximate number of parameters when publicly available; otherwise, we mark it as ``--''. 
Model categories are grouped by their primary editing or generation mechanism.
\end{tablenotes}
\end{threeparttable}
}
\end{table*}

\section{Judge Prompt and JSON Schema}
\label{app:judge-prompt}

This appendix provides the prompt design and structured output schema used by our MLLM-as-a-Judge evaluation protocol.
For each evaluation instance, the judge receives the source meme, the edited meme, the source emotion, the target emotion, and optional reference information, including the source caption, target caption, and editing instruction.
The judge is required to return a schema-constrained JSON object without any free-form text outside the JSON output.

\subsection{Judge Input}

Each evaluation sample is represented as:
\begin{equation}
x = \{I_{\mathrm{src}}, I_{\mathrm{edit}}, a_{\mathrm{src}}, a_{\mathrm{tgt}}, T_{\mathrm{src}}, T_{\mathrm{tgt}}, e\},
\end{equation}
where $I_{\mathrm{src}}$ and $I_{\mathrm{edit}}$ denote the source and edited meme images, $a_{\mathrm{src}}$ and $a_{\mathrm{tgt}}$ denote the source and target emotions, $T_{\mathrm{src}}$ and $T_{\mathrm{tgt}}$ denote optional source and target captions, and $e$ denotes the optional editing instruction.
The judge compares the edited meme against the source meme under the target emotion constraint.

\subsection{System Prompt}

The system prompt defines the role of the judge and enforces four global constraints:

\begin{promptbox}[title={System Prompt}]
You are an expert evaluator for the task: **Meme Reappraisal**.

You will be given a **SOURCE meme image**, an **EDITED meme image**, reference labels for **SOURCE emotion** and **TARGET emotion**, and optional reference captions and editing instructions.

Your job is to compare **SOURCE vs. EDITED** and output only a JSON object that strictly follows the provided JSON schema.

**Global constraints.**

1. **Content preservation.** 
The **EDITED meme** must preserve the same situation, entities, identity, and core meaning as the **SOURCE meme**.
Penalize outputs where the topic changes, new events are introduced, or the implied situation differs.

2. **Layout and structure preservation.**
If the **SOURCE meme** is single-panel, the **EDITED meme** should remain single-panel.
If the **SOURCE meme** is multi-panel or grid-based, the **EDITED meme** should preserve the panel structure, panel order, and caption regions.

Any mismatch should be marked as **layout inconsistency**.

3. **Target-gated emotion scoring.**
Emotion-alignment scores must first check whether the **EDITED meme** matches the given **TARGET emotion**. 
If the **EDITED meme** does not match the **TARGET emotion**, the emotion-alignment score must not exceed **2** on the **1--5** scale.

4. **Localized rationale.**
Each score must have its own rationale.

The rationale for each block should only justify that specific score or decision and should not be copied across blocks. 
Each rationale should mention target-emotion alignment, scenario preservation, and text-visual semantic alignment when applicable.

Return **JSON only** and do not add extra keys outside the schema.
\end{promptbox}

\subsection{User Prompt}
The user prompt provides the task instance and the criterion-specific rubrics.
The following template is used:

\begin{promptbox}[title={User Prompt}]
Task: Evaluate the **EDITED meme** for **Meme Reappraisal** given the **SOURCE meme** and **TARGET emotion**.

Allowed emotion labels:
`angry`, `sad`, `tense`, `bored`, `calm`, `happy`, `excited`, `relaxed`, and `other`.

SOURCE emotion: `{src_emotion}`

TARGET emotion: `{tgt_emotion}`

Optional references:
- SOURCE caption: `{src_caption}`
- TARGET caption: `{tgt_caption}`
- Editing instruction: `{edit_instruction}`

Evaluate the **EDITED meme** according to the following criteria:
visual assessment, text assessment, layout consistency, overall emotion classification, perceived emotion shift, and overall generation quality.

Each score should follow its own rubric and rationale.
\end{promptbox}

% \begin{quote}
% \small
% Task: Evaluate the EDITED meme for Meme Reappraisal given the SOURCE meme and TARGET emotion.

% Allowed emotion labels:
% \texttt{angry}, \texttt{sad}, \texttt{tense}, \texttt{bored}, \texttt{calm}, \texttt{happy}, \texttt{excited}, \texttt{relaxed}, and \texttt{other}.

% SOURCE emotion: \texttt{\{src\_emotion\}}

% TARGET emotion: \texttt{\{tgt\_emotion\}}

% Optional references:
% \begin{itemize}
% \item SOURCE caption: \texttt{\{src\_caption\}}
% \item TARGET caption: \texttt{\{tgt\_caption\}}
% \item Editing instruction: \texttt{\{edit\_instruction\}}
% \end{itemize}

% Evaluate the EDITED meme according to the following criteria:
% visual assessment, text assessment, layout consistency, overall emotion classification, perceived emotion shift, and overall generation quality.
% Each score should follow its own rubric and rationale.
% \end{quote}

The source and edited meme images are provided as image inputs.
In the Visual-First setting, the images are placed before the textual instruction.
In the Visual-Last setting, the textual instruction is placed before the images.

\subsection{Evaluation Rubric}

The judge evaluates six structured blocks, from which the eight fine-grained metrics are derived.

\paragraph{Visual Assessment.}
The judge scores:
\begin{itemize}
\item \texttt{generation\_quality\_1\_5}: visual cleanliness, artifact level, realism or style coherence, and readability of visual details.
\item \texttt{emotion\_accuracy\_1\_5}: whether visual affective cues match the target emotion, including facial expression, posture, lighting, color tone, and tension or relaxation.
\end{itemize}
If the visual affect does not match the target emotion, the visual emotion score is capped at 2.

\paragraph{Text Assessment.}
The judge scores:
\begin{itemize}
\item \texttt{generation\_quality\_1\_5}: text legibility, absence of garbling or cropping, formatting consistency, and natural meme-style placement.
\item \texttt{emotion\_accuracy\_1\_5}: whether the rewritten text expresses the target emotion while preserving the same scenario.
\end{itemize}
If the text does not convey the target emotion, the text emotion score is capped at 2.

\paragraph{Layout Consistency.}
The judge returns a Boolean value:
\begin{itemize}
\item \texttt{consistent}: true only if the edited meme preserves the source layout type, panel structure, panel order, and caption regions.
\end{itemize}

\paragraph{Overall Emotion Classification.}
The judge predicts the primary perceived emotion of the edited meme:
\begin{itemize}
\item \texttt{label}: one of the eight emotions or \texttt{other}.
\item \texttt{other\_emotion}: a specific emotion name if and only if \texttt{label} is \texttt{other}; otherwise null.
\end{itemize}

\paragraph{Perceived Emotion Shift.}
The judge scores:
\begin{itemize}
\item \texttt{magnitude\_1\_5}: the perceived strength of emotional change from the source meme to the edited meme.
\end{itemize}

\paragraph{Overall Generation Quality.}
The judge scores:
\begin{itemize}
\item \texttt{overall\_generation\_quality\_1\_5}: holistic quality considering target-emotion success, content preservation, layout preservation, and readability.
\end{itemize}

\subsection{JSON Output Schema}

The judge output follows the strict JSON schema below.
The order of the \texttt{rationale} field depends on the prompt configuration:
in Rationale-First settings, each block places \texttt{rationale} before scalar fields;
in Rationale-Last settings, each block places \texttt{rationale} after scalar fields.

\begin{quote}
\small
\begin{verbatim}
{
"visual_assessment": {
"rationale": string,
"generation_quality_1_5": integer in [1, 5],
"emotion_accuracy_1_5": integer in [1, 5]
},
"text_assessment": {
"rationale": string,
"generation_quality_1_5": integer in [1, 5],
"emotion_accuracy_1_5": integer in [1, 5]
},
"layout_consistency": {
"rationale": string,
"consistent": boolean
},
"overall_emotion_classification": {
"rationale": string,
"label": one of [
"angry", "sad", "tense", "bored",
"calm", "happy", "excited", "relaxed", "other"
],
"other_emotion": string or null
},
"perceived_emotion_shift": {
"rationale": string,
"magnitude_1_5": integer in [1, 5]
},
"overall_generation_quality": {
"rationale": string,
"overall_generation_quality_1_5": integer in [1, 5]
}
}
\end{verbatim}
\end{quote}

We additionally apply post-validation to ensure that \texttt{other\_emotion} is non-empty when \texttt{label} is \texttt{other}, and null otherwise.

\subsection{Metric Mapping}

The JSON fields are mapped to the eight metrics used in the main paper as follows:

\begin{table}[t]
\centering
\tabcolsep 4pt
\resizebox{1.0\linewidth}{!}{
\begin{tabular}{|l|l|}
\hline
Metric & JSON Field \\
\hline
VGQ & \texttt{visual\_assessment.generation\_quality\_1\_5} \\
VEA & \texttt{visual\_assessment.emotion\_accuracy\_1\_5} \\
TGQ & \texttt{text\_assessment.generation\_quality\_1\_5} \\
TEA & \texttt{text\_assessment.emotion\_accuracy\_1\_5} \\
LC  & \texttt{layout\_consistency.consistent} \\
HGQ & \makecell[l]{\texttt{overall\_generation\_quality.}\\
\texttt{overall\_generation\_quality\_1\_5}} \\
PPE & \makecell[l]{\texttt{overall\_emotion\_classification.}\\
\texttt{label}} \\
PES & \texttt{perceived\_emotion\_shift.magnitude\_1\_5} \\
\hline
\end{tabular}}
\caption{Mapping between judge JSON fields and evaluation metrics.}
\label{tab:judge-metric-mapping}
\end{table}

Scalar 1--5 scores are normalized to $[0,1]$ for aggregated scoring.
For LC, \texttt{true} is mapped to 1 and \texttt{false} is mapped to 0.
For PPE, we compute the target emotion hit indicator by checking whether the predicted label matches the target emotion.

\subsection{Prompt Configuration Ablation}

We evaluate four prompt configurations by varying two orthogonal factors:
\begin{itemize}
\item \textbf{Evidence order}: Visual-First (VF) places image evidence before textual instructions, while Visual-Last (VL) places image evidence after textual instructions.
\item \textbf{Reasoning order}: Rationale-First (RF) places the rationale before scalar scores in each JSON block, while Rationale-Last (RL) places the rationale after scalar scores.
\end{itemize}

Their Cartesian product yields four configurations:
RF-VF, RF-VL, RL-VF, and RL-VL.
The main experiments use the configuration that best aligns with human expert judgments.

\section{Additional Qualitative Results}
\label{app:qualitative-results}

Fig.~\ref{fig:appendex_usecase} provides the full model-wise qualitative comparison for the examples in the main text.
Due to space constraints, the main paper shows only a subset of representative outputs, while the appendix includes results for all evaluated models.

\section{Discussion}
\label{sec:discussion}

Our results suggest that Emotional Reappraisal is a distinct capability beyond meme understanding, meme generation, and generic emotion-controllable editing.
Compared with meme-understanding and safety-oriented benchmarks such as Hateful Memes and GOAT-Bench~\cite{kiela2020hateful,lin2026goat}, MER reveals a gap between recognizing affective meaning and transforming it under structural constraints.
Identifying the emotion of a meme is substantially easier than reinterpreting it while preserving its visual layout, embedded text, and semantic context.

Relative to meme generation methods such as MemeCap, XMeCap, and IterMeme~\cite{hwang2023memecap,chen2024xmecap,cai2025itermeme}, MER introduces an invariance requirement: the edited meme should preserve the original identity, layout, and core situation while changing its affective interpretation.
This shifts the problem from open-ended generation to constrained transformation.
Empirically, models that are strong in image synthesis, layout preservation, or text rewriting alone do not reliably achieve successful reappraisal when these objectives must be satisfied jointly.

MER also extends prior work on emotion-controllable generation~\cite{zhang2024emotion,yang2025emoctrl,yang2024emogen,brooks2023instructpix2pix}.
While existing methods demonstrate affect control in generic settings, our results show that emotion manipulation becomes substantially harder when affect is encoded through tightly coupled visual content, embedded text, layout structure, and meme conventions.
This multimodal interdependence creates a structural-affective trade-off: current models often either preserve meme structure without changing emotional framing, or modify affective cues while disrupting layout, text rendering, or semantic identity.

\subsection{Application Scenario}
\label{sec:application-scenario}

\begin{figure}[!t]
\centering
\includegraphics[width=\columnwidth]{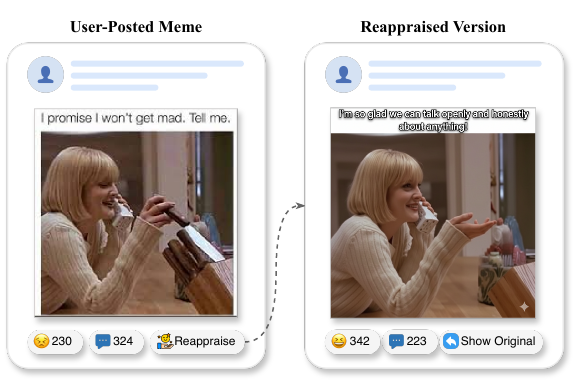}
\caption{Example application scenario of Meme Reappraisal. 
A user-shared negative meme (left) is reinterpreted into a positive version (right) while preserving the original situation and meme structure. 
This illustrates its potential for constructive reframing in emotionally aware online communities.}
\label{fig:usecase_meme_reappraisal}
\vspace{-1.5em}
\end{figure}
Fig.~\ref{fig:usecase_meme_reappraisal} illustrates a potential application scenario of MER.
A user-posted negatively framed meme can be reinterpreted into a more constructive version while keeping the same situation and meme structure.
This example highlights the practical motivation for studying structure-preserving affect transformation, but also underscores the need for careful evaluation: effective reappraisal should not simply make a meme ``positive'', but should preserve what the meme is about while changing how its situation is emotionally framed.

Overall, these findings indicate that structure-preserving affect transformation constitutes a distinct evaluation axis for multimodal systems.
MER-Bench, therefore, complements existing benchmarks by testing whether models can coordinate visual editing, text rewriting, layout preservation, and affective reinterpretation within a single constrained multimodal task.

\clearpage


\end{document}